\documentclass{new_tlp}
\usepackage{mathptmx}
\usepackage{quoting}

\usepackage{thm-restate}
\declaretheorem[numberwithin=section]{theorem}
\declaretheorem[numberwithin=section]{proposition}
\declaretheorem[numberwithin=section]{definition}
\declaretheorem[numberwithin=section]{example}
\declaretheorem[numbered=yes]{lemma }

\usepackage{times}
\usepackage{soul}
\usepackage{url}
\usepackage{hyperref}
\usepackage[utf8]{inputenc}
\usepackage[small]{caption}
\usepackage{floatrow}
\usepackage{graphicx}
\usepackage{amsmath}
\usepackage{booktabs}
\usepackage{algorithm}
\usepackage{algorithmic}
\usepackage{color}
\usepackage{stackengine}

\usepackage[T1]{fontenc}
\usepackage{courier}

\usepackage[label font=bf]{subfig}
\usepackage{listings}
\lstdefinestyle{asp-style}{
	language=Prolog,
	frame=lines,
	keywordstyle=\linespread{1.1}\scriptsize\ttfamily,
	basicstyle=\linespread{1.1}\scriptsize\ttfamily,
	breaklines=true,
}
\usepackage[english]{babel}
\usepackage{xspace}
\usepackage{amssymb}
\usepackage{todonotes}
\usepackage{pdfsync}
\usepackage{paralist}
\usepackage{srcltx}
\usepackage[acronym,nonumberlist,nohypertypes={acronym}]{glossaries}

\usepackage{tikz}
\usepackage{subfig}
\usetikzlibrary{external}
\usepackage{pgfplots}
\usepgfplotslibrary{external}
\usetikzlibrary{plotmarks}
\usetikzlibrary{pgfplots.groupplots}
\pgfplotsset{
	filter discard warning=false % remove notes while compiling
	, legend cell align=left
	, minor grid style={loosely dotted, lightgray}
	, major grid style={loosely dashed, lightgray}
}

\makeatletter
\renewcommand{\todo}[2][]{\tikzexternaldisable\@todo[#1]{#2}\tikzexternalenable}
\makeatother

\newacronym{dlp}{\textsc{DLP}}{{Disjunctive Logic Programming}}
\newacronym{asp}{\textsc{ASP}}{{Answer Set Programming}}

\def\derives{\ensuremath{:\!-}}

\def\HA{Heads\xspace}

\newcommand{\grnd}{\ensuremath{grnd}\xspace}

\newcommand{\X}{{\bf X}}

\newcommand{\UF}{\ensuremath{U\!F}}

\newcommand{\veel}{ \ |\ }

\newcommand{\nop}[1]{}

\def\embeds{\vdash}
\def\nembeds{\nvdash}

\newcommand{\EP}[1]{\ensuremath{{\cal E}_{#1}}}

\def\UF{\ensuremath{U\!F}}

\def\EPS{{{\cal E}\!\!{\cal S}}}

\def\pacman{{Pac-Man}\xspace}

\newcommand\quo[1]{``#1''}

\newcommand{\inst}{{\ensuremath Inst}}
\newcommand{\instt}[2]{{\ensuremath{{ Inst}({#1},{#2})}}}
\newcommand{\insttinf}[2]{{\ensuremath{{Inst}({#1},{#2})^\infty}}}
\newcommand{\insttk}[3]{{\ensuremath{{Inst}({#1},{#2})^{#3}}}}

\newcommand{\sysfont}{\textit}
\newcommand{\dlv}{\sysfont{DLV}\xspace}
\newcommand{\idlv}{{{\small $\cal I$}-}\dlv}
\newcommand{\system}{\idlv-incr\xspace}
\newcommand{\systemnoincr}{\idlv-no-incr\xspace}

    \title[Incremental Answer Set Programming with Overgrounding]
        {Incremental Answer Set Programming \\ with Overgrounding}

    \author[Calimeri, Ianni, Pacenza, Perri and Zangari]{
        Francesco Calimeri,
        Giovambattista Ianni,
        Francesco Pacenza,
        Simona Perri,
        Jessica Zangari
        \\
        Department of Mathematics and Computer Science, University of Calabria, Italy \\
        \email{{\em lastname}@mat.unical.it} - \url{https://www.mat.unical.it}
    }

\jdate{}
\pubyear{}
\submitted{May 15th, 2019}
\revised{June 19th, 2019}
\accepted{July 17th, 2019}

\begin{document}

\label{firstpage}

\maketitle

\begin{abstract}
 Repeated executions of reasoning tasks for varying inputs are necessary in many applicative settings, such as stream reasoning.
In this context, we propose an incremental grounding approach for the answer set semantics.
We focus on the possibility of generating incrementally larger ground logic programs equivalent to a given non-ground one; so called {\em overgrounded programs} can be reused in combination with deliberately many different sets of inputs.
Updating overgrounded programs requires a small effort,
thus making the instantiation of logic programs considerably faster when grounding is repeated on a series of inputs similar to each other.
Notably, the proposed approach works \quo{under the hood}, relieving designers of logic programs from controlling technical aspects of grounding engines and answer set systems.
In this work we present the theoretical basis of the proposed incremental grounding technique, we illustrate the consequent repeated evaluation strategy and report about our experiments. This paper is under consideration in Theory and Practice of Logic Programming (TPLP).
\end{abstract}


%\todo{SITO}

%%%%%%%%%%%%%%%%%%%%%%%%%%%%%%%%%%%%%%%%%%%%%%%%%%%%%%%%%%%%%%%%%%%%%%%%%%%
\section{Introduction}\label{sec:intro}
The practice of attributing meaning to quantified logical sentences relying on equivalent propositional versions thereof dates back to the historical work of Jacques Herbrand~\cite{herbrand1930recherches}.
Later, at the end of the past century, in the trail of such practice, {\em ground programs} have been used as the operational basis for computing the semantics of logic programs in the context of the well-founded semantics~\cite{vangelder-etal:1991} and of the answer set semantics~\cite{gelf-lifs-1991,DBLP:conf/rweb/EiterIK09}.
Indeed, the typical structure of an Answer Set Programming (ASP) system implements its semantics by relying on a grounding module that takes in input a non-ground logic program $P$ and produces an equivalent propositional theory $gr(P)$; the grounder is coupled with a subsequent solver module that applies proper resolution techniques on $gr(P)$ for computing the actual semantics in form of {\em answer sets}~\cite{DBLP:journals/aim/KaufmannLPS16}.
Steps similar to grounding are also taken when high-level input specifications are transformed in low-level constraint sets, or propositional SAT theories, such as when elaborating MiniZinc scripts~\cite{DBLP:conf/cp/NethercoteSBBDT07}.
%\todo{verificare dove e come riportare minizinc, cosi' sembra spezzare troppo.}

The cost of generating a propositional theory can be significant both in terms of time and space; that is why the grounding phase cannot be overlooked just as a preprocessing stage.
There are a large number of both benchmark and practical application settings in which the grounding step is prominent in terms of used resources~\cite{DBLP:journals/jair/GebserMR17}.
Note also that, as soon as variable programs are given in input, the produced ground program is potentially of exponential size with respect to the input program\footnote{Strictly speaking, if variable arity and variable rule/program lengths are allowed, the program complexity of the decisional problem associated to the grounding of Datalog programs is complete for EXPTIME (Th. 4.5 of \cite{DBLP:journals/csur/DantsinEGV01}).}.
As the popularity of ASP and its use increased, the scientific community worked on both theoretical and technical sides and produced a number of optimization techniques for reducing space and time costs of the grounding step~\cite{DBLP:conf/iclp/CalimeriCIL08,DBLP:conf/lpnmr/GebserKKS11,DBLP:journals/ai/AlvianoFGL12,DBLP:journals/ia/CalimeriFPZ17}, or to blend it within the answer set search phase to some extent~\cite{DBLP:journals/fuin/PaluDPR09,DBLP:conf/jelia/Dao-TranEFWW12,deneck-incrground-2012,DBLP:conf/lpnmr/Weinzierl17,DBLP:journals/tplp/LefevreBSG17}.

It must be observed that some contexts require to reason over data streams or via multiple, subsequent \quo{shots}~\cite{DBLP:journals/tplp/BeckEB17,DBLP:journals/tplp/GebserKKS19}; in such cases, where input data changes with time, instantiation can be quite a big time bottleneck, as not only it must be repeatedly computed on slightly different and dynamically changing input data, but also only short time windows are allowed between a \quo{shot} and another.
%reasoning or
%Take, for instance, those contexts in which a reasoning task must be repeatedly performed on input data changing over time, such as the ones in which
%
%such as stream reasoning and multi-shot evaluation~\cite{DBLP:journals/tplp/BeckEB17,DBLP:journals/tplp/GebserKKS19}.
%, grounding is particularly relevant, and can be quite a big time bottleneck.
%
%In such contexts, instantiation can be quite a big time bottleneck, as it must be repeatedly computed on slightly different and dynamically changing input data while only a short time window is allowed between a ``shot'' and another.
%
For instance, when dealing with real-time videogames, artificial players are allowed a very limited time per each decision.
In competitions like the known GVGAI (General Video Game AI)~\cite{DBLP:conf/aaai/LiebanaSTSL16}, this limit is just $40$ milliseconds, which is quite challenging even for state-of-the-art answer set systems.
In such cases, the solving step is usually easy from the computational point of view; this makes the grounding step prominent in terms of optimization requirements.

A remarkable contribution has been introduced with the {\em iclingo} and {\em oclingo} systems~\cite{DBLP:conf/lpnmr/GebserKKS11,DBLP:journals/tplp/GebserKKS19}.
In {\em oclingo}, later generalized and integrated in latest version of the monolithic {\em clingo} system, a designer can procedurally control which parts of a logic program must be kept constant between two consecutive shots, thus allowing the caching of ground subprograms and of partial answer sets.
Also, it is possible to control which parts of a program are subject to incremental evaluation with respect to an iteratively increasing integer parameter.% $t$.
Nevertheless, procedural controllability is not desirable in many development scenarios, especially because it requires a non-negligible knowledge of solver-specific internal algorithms.
Let us cite, again, the videogame industry; while developing, designers look for easy and off-the-shelf solutions, and often do not have knowledge of declarative logic programming at all.
Among other notable works, the Ticker incremental evaluation system~\cite{DBLP:journals/tplp/BeckEB17} implements LARS, a stream reasoning formal framework with ASP-like semantics.
LARS allows omni-comprehensive, yet demanding, temporal data management features which makes implementation more difficult and complex.

In this work, we focus on the usage of the pure ASP semantics in a repeated evaluation setting.
In particular, we aim at closing the abovementioned gaps in the current state of the art by investigating the possibility of generating incrementally larger ground programs for a fixed non-ground logic program.
These ground programs can be reused in combination with deliberately many different sets of inputs; also the knowledge designer is relieved from the burden of manually controlling the computational procedure.
%with respect to the solutions mentioned above, yet easing the burden of taking care of it: indeed, the overgrounding process and the multi-shot machinery can be completely transparent to the user.
%\todo{verificare che i claim siano ragionevoli. GB: riscritta frase.}

\medskip

The contributions of this paper are summarized as follows.
\begin{inparaitem}
\item[$(i)$:\ ]
    We characterize a class of ground logic programs equivalent to the theoretical instantiation, called {\em embeddings} or {\em embedding programs}; they allow us to introduce a model-theoretic-like notion of ground programs with some desirable properties,
    and make dealing with formal properties of ground programs cleaner.
    {\em Overgrounded programs} are series of embedding programs growing monotonically that have both theoretical and practical impact: for instance, overgrounded programs can be easily generalized to other semantics for logic programming, such as the well-founded semantics; moreover, their incremental growth allows for easily implementing caching policies in many practical scenarios.
\item[$(ii)$:\ ]
    We propose an incremental grounding strategy, allowing to reuse previously grounded programs in consecutive evaluation shots.
    In particular, while dealing with a specific reasoning task, we maintain a stored ground logic program which grows monotonically from one shot to another; such an {\em overgrounded program} becomes more and more general while moving from a shot to the next, increasingly adding potentially useful rules.
    Importantly, intermediate subsequent updates to the ground program are considerably less time-consuming: our technique allows, in a sense, to trade memory for time.
\item[$(iii)$:\ ]
    We report about the experimental activities we conducted, aimed at assessing the practical impact of our approach; results show that it pays off in terms of performance, by knocking down grounding times and keeping solving times within more than reasonable bounds.
    We expect this setting to be particularly favourable when non-ground input programs may contain grounding-intensive rules, as for instance, in the case of declaratively programmed robots or videogame agents.
\item[$(iv)$:\ ]
    We relieve designers of logic programs from two specific burdens.
    First, there is no need for procedural control over the incremental evaluation process, as features of the herein proposed incremental framework are almost transparent to designers.
    Second, there is no need to worry about which parts of logic programs might be more grounding-intensive; also highly general code, even non-optimized, can benefit from overgrounding.
    This allows to focus on representing knowledge in a single-encoding/multiple-inputs setting, thus preserving some desirable properties of knowledge representation and reasoning via ASP, namely declarativity and easiness of modelling.
\end{inparaitem}

\medskip

In the following, after overviewing our approach and briefly presenting some preliminaries, we illustrate the theoretical basis our grounding strategy relies on.
We then illustrate our framework and report about our experiments; we eventually discuss related work before drawing some final considerations. 
%For space reasons, proofs are reported in \ref{appendix}.
For space reasons, proofs are reported in the supplementary material attached to this paper at the TPLP archives.

%%%%%%%%%%%%%%%%%%%%%%%%%%%%%%%%%%%%%%%%%%%%%%%%%%%%%%%%%%%%%%%%%%%%%%%%%%%
\section{Overgrounding: an overview}
A canonical ASP system works by first {\em instantiating} a non-ground logic program $P$ over input facts $F$, obtaining a propositional logic program $gr(P,F)$, and then computing the corresponding intended models, i.e., the answer sets $AS(gr(P,F))$.
Importantly, systems build $gr(P,F)$ as a significantly smaller and refined version of the theoretical instantiation, defined via the Herbrand base
(see, e.g.,~\cite{DBLP:conf/iclp/CalimeriCIL08}), but preserve semantics, i.e., $AS(gr(P,F)) = AS(P \cup F)$.
The choice of the instantiation function $gr$ impacts on both computing time and on the size of the obtained instantiated program.
$gr$ usually maintains a set $PT$ of ``possibly true'' atoms, initialized as $PT = F$; then, $PT$ is iteratively incremented and used for instantiating only ``potentially useful'' rules, up to a fixpoint.
Strategies for decomposing programs and for rewriting, simplifying and eliminating redundant rules can be of great help in controlling the size of the final instantiation\footnote{For an overview of grounding optimization techniques the reader can refer to~\cite{DBLP:conf/lpnmr/GebserKKS11,DBLP:journals/ia/CalimeriFPZ17,cali-perr-zang-TPLP-optimizing}.}.

\begin{example}\label{exmp:overgrouding}

We show the overgrounding approach with a simple example.
Let us consider the
%{\em fixed}
program
$P_0$:
\begin{center}
\begin{tabular}{lll}
$r(X,Y) \ \derives \ e(X,Y),\ not\ ab(X).$ & \hspace{5mm} &
$r(X,Z) \veel s(X,Z) \ \derives \ e(X,Y),\ r(Y,Z).$
\end{tabular}
\end{center}

When taking the set of facts $F_1 = \{ e(c,a),\ e(a,b) \}$ into account, there are several ways for building a tailored instantiation of $P_0 \cup F_1$.
For instance, one can simply assume $F_1$ as the initial set of ``possibly true'' facts, then generate new rules and new possibly true facts by iterating through positive head-body dependencies, obtaining the ground program $G_1$:

\begin{center}
\begin{tabular}{lll}
$r_1 : r(a,b) \ \derives \ e(a,b),\ not\ ab(a).$ & \hspace{5mm} &
$r_2 : r(c,b)  \veel s(c,b) \ \derives \ e(c,a),\ r(a,b).$ \\
$r_3 : r(c,a) \ \derives \ e(c,a),\ not\ ab(c).$
\end{tabular}
\end{center}

A more ``aggressive'' grounding strategy could also cut or simplify rules.
For instance, one can use a simplification strategy which eliminates negative literals that are identified as definitely true.
In the case above, it is easy to see that $ab(a)$ and $ab(c)$ have no chance of being true; hence, removing $not\ ab(a)$ and $not\ ab(c)$ from the rule bodies leads to the generation of $G'_1$:

\begin{center}
\begin{tabular}{lll}
$r(a,b) \ \derives \ e(a,b).$ & \hspace{5mm} &
$r(c,b)  \veel s(c,b) \ \derives \ e(c,a),\ r(a,b).$ \\
$r(c,a) \ \derives \ e(c,a).$
\end{tabular}
\end{center}

Nevertheless, $G'_1$ can be seen as less general, less re-usable than $G_1$, as it cannot be easily extended to a program which is equivalent to $P_0$ with respect to larger classes of input facts.
Let us assume that, at some point, a subsequent run requires $P_0$ to be evaluated over a different set of input facts $F_2 = \{ e(c,a),\ e(a,d),$\ $ab(c) \}$. Note that, with respect to $F_1$, $F_2$ features the additions $F^+ = \{ e(a,d), ab(c) \}$ and the deletions $F^- = \{ e(a,b) \}$.
Nonetheless, differently from $G'_1$, $G_1$ can be easily incrementally updated by adding $F^+$ to the set of possibly true facts, yet preserving semantics.
More precisely, one can ground $P_0$ with respect to the new set of possibly true facts $F_1 \cup F_2$; then, this new information can be propagated and only some additional rules $\Delta G_1 = \{ r_4, r_5 \}$, must be added, thus obtaining $G_2$:

\begin{center}
\begin{tabular}{lll}
$r_1 : r(a,b) \ \derives \ e(a,b),\ not\ ab(a).$ & \hspace{5mm} &
$r_2 : r(c,b)  \veel s(c,b) \ \derives \ e(c,a),\ r(a,b).$ \\
$r_3 : r(c,a) \ \derives \ e(c,a), not\ ab(c).$ & \hspace{5mm} &
${\bf r_4 : r(c,d)  \veel s(c,d) \ \derives \ e(c,a),\ r(a,d).}$ \\
${\bf r_5 : r(a,d) \ \derives \ e(a,d),\ not\ ab(a).}$
\end{tabular}
\end{center}

We have now that $G_2$ is equivalent to $P_0$, both when evaluated over $F_1$ and when evaluated over $F_2$ as input facts, although only different portions of the whole set of rules in $G_2$ can be considered as relevant when considering $F_1$ or $F_2$ as inputs, respectively.
On the other hand, $G'_1$ cannot be easily incremented with new rules so to be ``compatible'' with $F_2$, because the rule $r(c,a) \ \derives \ e(c,a)$ would cause wrong inferences.
Even more interestingly, if a third {\em shot} of reasoning is requested over input facts $F_3 = \{ e(a,d),$\ $e(c,a),$\ $e(a,b)\}$, we notice that $G_2$ does not require any further incremental update, as possibly true facts remain unchanged (i.e., $F_1 \cup F_2$ = $F_1 \cup F_2 \cup F_3$), i.e. the set $\Delta G_2$ containing new incremental additions to $G_2$ would be empty.
\end{example}
It turns out that an instantiation strategy like the one producing $G_1$ and then $G_2$ should comply with specific properties which allow to define an incremental grounding strategy.
Such properties are illustrated and used in the remainder of the paper.

%%%%%%%%%%%%%%%%%%%%%%%%%%%%%%%%%%%%%%%%%%%%%%%%%%%%%%%%%%%%%%%%%%%%%%%%%%%
\section{Preliminaries}\label{sec:preliminaries}
Throughout the paper we will assume to deal with programs under the answer set semantics.
%As a flexible language for declarative problem solving, it allows the user to refrain from providing an algorithm for solving the problem at hand: it is sufficient to specify the properties a desired solution must (or must preferably) have in the form of an executable specification and run this latter to obtain the so called {\em answer sets}.
%
A program $P$ is a set of rules. A rule $r$ is of the form:
$\alpha_1 \veel \alpha_2 \veel \dots \veel \alpha_k$ :- $\beta_1,\ \dots,\ \beta_n,$\ $not$ $\beta_{n+1},\ \dots,$\ $not$ $\beta_m$,\ \
where $n, m, k \geqslant 0$.
$\alpha_1, \dots, \alpha_k$ and $\beta_1, \dots, \beta_m$ are called {\em atoms}.
An atom is of the form $p(\X)$, for $p$ a predicate name and $\X$ a list of variable names and constants.
The {\em head} of $r$ is defined as $H(r) = \{\alpha_1,\ \dots,\ \alpha_k\}$; if $H(r) = \emptyset$ then $r$ is said to be a {\em constraint}.
The {\em positive body} of $r$ is defined as $B^+(r) = \{\beta_1, \dots, \beta_n\}$, while the {\em negative body} as $B^-(r) = $ $\{not$ $\beta_{n+1}, \dots, $ $not$ $\beta_m\}$.
The {\em body} of $r$ is defined as $B(r) = B^+(r) \cup B^-(r)$; if $B(r) = \emptyset$ and $\|H(r)\| = 1$ then $r$ is referred to as a {\em fact}.
The set of all head atoms in $P$ is denoted by $\HA(P) = \bigcup_{r \in P} H(r)$.
As usual, we deal with ``safe'' logic programs, i.e., for any non-ground rule $r \in P$, and for any variable $X$ appearing in $R$, there is at least an atom $p \in B^+(r)$ mentioning $X$.

A program (resp. a rule, a literal, a term) is said to be {\em ground} if it contains no variables.
%
%\todo{GB: Verificare quanto diciamo sull'universo delle costanti}
%
In the following, we will assume to deal with a fixed {\em Herbrand Universe} consisting of a finite set of constants $U$, and to deal with programs that do not contain facts, as these are separately given in form of sets of ground atoms: given a program $P$ and a set of facts $F$, both $P$ and $F$ will only feature constant symbols appearing in $U$.
%
%Given a logic program $P$ and a set of facts $F$, the {\em Herbrand universe} of $P$ and $F$, denoted by $U_{P,F}$, consists of all
%ground terms that can be built combining constants appearing in $P$ or in $F$.
The {\em Herbrand base} of a program $P$, denoted by $B_{P}$, is the set of all ground atoms obtainable from the atoms of $P$ by replacing variables with elements from $U$.
A {\em substitution} for a rule $r \in P$ is a mapping from the set of variables of $r$ to the set $U$ of ground terms.
A {\em ground instance} of a rule $r$ is obtained by applying a substitution to $r$.

Given a logic program $P$, the {\em theoretical instantiation (grounding)} of $P$, denoted as $\grnd(P)$, is defined as the set of all ground instances of rules in $P$.
We assume the reader is familiar with the notions of interpretation and model as subsets of $B_P$, and with the usual notation in the literature; in particular, when an interpretation $I$ models an element $e$ (resp. an atom, a body, a head, a rule) this is denoted by $I \models e$.
Given $P$, and also a set of facts $F$, an answer set $A$ of $P \cup F$ is a subset of $B_P$,
defined as the minimal model of the so-called FLP reduct $(\grnd(P) \cup F)^A$ of $\grnd(P) \cup F$~\cite{DBLP:conf/jelia/FaberLP04}.
We denote the set of all answer sets of $P\cup F$ as $AS(P \cup F)$.

%$e(P) = ( KB \cup F \cup F^+ ) \setminus F^- $ \\

%NOTA: valutare se F pu\`{o} essere suddiviso in elementi costanti e variabili\\
%\vspace{2mm}

%Answer Set parziale: A $\in$ AS(P), $\widetilde{A}$ \`{e} un sottoinsieme di A\\
%\vspace{2mm}

%Stato parziale = $AS \cup AS(P) \cup \widetilde{AS}$ where $\widetilde{AS}$ is a set of partial answer sets.\\
%\vspace{2mm}

%problem: it is given a program P, a partial state AP, and an evolution e(P). Compute AS(e(P)), possibly
%reusing AP.

%\medskip

It is possible to compute a ground program equivalent to $\grnd(P)\cup F$ by means of the operator defined next. In this section we will follow~\cite{DBLP:conf/iclp/CalimeriCIL08}: note, however, that the instantiation operator defined here is slightly differently formalized, in that we purposely keep facts explicitly separated from rules.

\begin{definition}\label{def:instOperator} \rm
{\em [cf.~\cite{DBLP:conf/iclp/CalimeriCIL08}].}\ Given a program $P$ and a set of ground atoms $S$, we define the operator
$\instt{P}{S}$ as $\instt{P}{S} = \{ r \in \grnd(P)\ s.t.\ B^+(r) \subseteq S \}$. With a slight abuse of notation, for a set $R$ of ground rules we define $\instt{P}{R}$ as $\instt{P}{\HA(R)}$. $\insttk{P}{F}{k}$ is defined as the $k$-th element of the sequence $\insttk{P}{F}{0}$ = $\instt{P}{\emptyset \cup F}$, $\dots$, $\insttk{P}{F}{k}$ = $\instt{P}{\insttk{P}{F}{k-1}\cup F}$.
\end{definition}

Intuitively, the notion above formalizes the idea of selecting,
among all ground instances of rules in $P$, those {\em supported} by a given set $S$ or by the heads of a given set of ground rules $R$. Given that the sequence above is defined by a monotonic operator it converges to its least fixpoint.
%
%By iteratively applying the $\inst$ operator, the following sequence can be defined.
%
%\begin{definition}\label{def:instSequence}\rm
%For a program $P$ and set of facts $F$, $\insttk{P}{F}{k}$ is $k$-th element of the sequence $\insttk{P}{F}{0}$ =
%$\instt{P}{\emptyset \cup F}$, $\dots$, $\insttk{P}{F}{k}$ = $\instt{P}{\insttk{P}{F}{k-1}\cup F}$.
%\end{definition}
\begin{proposition}\rm\label{prop:instconvergence}\rm
{\em [cf.~\cite{DBLP:conf/iclp/CalimeriCIL08}].}
Given a program $P$ and a set of facts $F$, the sequence $\insttk{P}{F}{k}$,  $k \geq 0$ converges to its least fixpoint $\insttinf{P}{F}$.
\end{proposition}

Interestingly, given a program $P$ and a set of facts $F$, the fixpoint above can be useful for computing a ground program that is equivalent to $\grnd(P) \cup F$.
\begin{theorem}\label{theo:instInfCorretto}{\em [cf.~\cite{DBLP:conf/iclp/CalimeriCIL08}]}
For a program $P$ and a set of facts $F$, $AS(P \cup F )$ $=$ $AS(\insttinf{P}{F}\cup F)$.
\end{theorem}

The following theorem extends Theorem 3.1 of~\cite{DBLP:journals/mlcs/Fages94} to the case of logic programs allowing disjunction in the heads; the proof descends from the notion of computation and Theorem~$6.22$ as given in~\cite{DBLP:journals/iandc/LeoneRS97}; for space reasons, we refrain from reporting the full definition here and refer the reader to the cited work.

\begin{theorem}\label{theo:therearestages}\rm
For a given program $P$, a set of facts $F$ and an answer set $A$ for $P$, we can assign to each atom $a \in A$ an integer value $stage(a) =$ $ i$ so that $stage$ encodes a strict well-founded partial order over all atoms in $A$, in such a way that there exists a rule $r \in \grnd(P) \cup F$ with $(i)$: $a \in H(r)$, $(ii)$: $A \models B(r)$ and $(iii)$: for any atom $b \in B^+(r)$, $stage(b) < stage(a)$.

\end{theorem}
%\begin{proof}
%It follows from the notion of computation and theorem 6.22 as given in~\cite{DBLP:journals/iandc/LeoneRS97}.
%\end{proof}

%%%%%%%%%%%%%%%%%%%%%%%%%%%%%%%%%%%%%%%%%%%%%%%%%%%%%%%%%%%%%%%%%%%%%%%%%%%
\section{Embedding Programs}\label{sec:embedding}
We introduce a declarative characterization of a class of equivalent ground programs, called {\em embeddings}.
This notion is useful for proving, given a program $P$ and a set of facts $F$, whether a ground program $G$ having certain features is equivalent to $P \cup F$, i.e., $G$ is a ``correct grounding'' that ``embeds'' $P\cup F$.
In a sense, embeddings relate to partial instantiations generated by the $\inst$ operator, like Herbrand models relate to supported interpretations generated by the immediate consequence operator.
An illustrative comparison between models and embedding programs is reported in Table~\ref{fig:embcomparison}.

\begin{definition}\label{def:embedding}\rm
[{\em Embeddings.}]\ For a program $P$, a set of facts $F$, a set of ground rules $R \subseteq (\grnd(P)\cup F)$, and a rule $r \in (\grnd(P) \cup F)$, we say that:
\begin{compactitem}
\item
    $R$ embeds the body of $r$, denoted $R\ \embeds_b r$, if $\forall a \in B^+(r)$ $\exists r' \in R$ s.t. $a \in H(r')$;
    %\item[(ii)]
%        $\forall b \in B^-(r)$, where $b = not\ a$,\ $\nexists r' \in \pri[2]$ s.t. $a \in H(r')$ and $r' \in Facts(\pri[2])$;
%    \end{itemize}
\item
    $R$ embeds the head of $r$, denoted $R\ \embeds_h r$, if $r \in R$.
\item
    $R$ embeds $r$, denoted $R\ \embeds r$, if either: $(i)$ $R\ \nembeds_b r$, or $(ii)$ $R\ \embeds_h r$.
\end{compactitem}
%\end{definition}
%
%\begin{definition}\label{def:EM}[{\em Embedding Programs.}]\
%Given a logic program $P$ and a set of input facts $F$,
%\begin{definition}\label{def:EM}[{\em Embedding Programs.}]\

\noindent Given a logic program $P$ and a set of input facts $F$, a set of ground rules $\EP{} \subseteq \grnd(P)\cup F$ is an {\em embedding program} for $P \cup F$, if $\forall r \in \grnd(P)\cup F$, $\EP{} \embeds r$.
\end{definition}

\begin{example}\label{exmp:embedding}
Let us consider the program $P_0$ of Example~\ref{exmp:overgrouding} along with the set of facts $F=\{e(a,b),e(b,b)\}$. The ground program below represents $F \cup grnd(P_0)$.
 %i.e. the theoretical grounding of $P_0$ over $F_1$.

\begin{center}
\begin{tabular}{lll}
$r_1 : r(a,a) \ \derives \ e(a,a),\ not\ ab(a).$ & \hspace{5mm} &
$r_2 : r(a,b) \ \derives \ e(a,b),\ not\ ab(a).$ \\
$r_3 : r(b,a) \ \derives \ e(b,a),\ not\ ab(b).$ & \hspace{5mm} &
$r_4 : r(b,b) \ \derives \ e(b,b),\ not\ ab(b).$ \\
$r_5 : r(a,a) \veel s(a,a) \ \derives \ e(a,a),\ r(a,a).$ & \hspace{5mm} &
$r_6 : r(a,a) \veel s(a,a) \ \derives \ e(a,b),\ r(b,a).$ \\
$r_7 : r(a,b) \veel s(a,b) \ \derives \ e(a,a),\ r(a,b).$ & \hspace{5mm} &
$r_8 : r(a,b) \veel s(a,b) \ \derives \ e(a,b),\ r(b,b).$ \\
$r_9 : r(b,a) \veel s(b,a) \ \derives \ e(b,a),\ r(a,a).$ & \hspace{5mm} &
$r_{10} : r(b,a) \veel s(b,a) \ \derives \ e(b,b),\ r(b,a).$ \\
$r_{11} : r(b,b) \veel s(b,b) \ \derives \ e(b,a),\ r(a,b).$ & \hspace{5mm} &
$r_{12} : r(b,b) \veel s(b,b) \ \derives \ e(b,b),\ r(b,b).$ \\
$r_{13} : e(a,b).$ & \hspace{5mm} &
$r_{14} : e(b,b).$
\end{tabular}
\end{center}

By definition, $grnd(P_0) \cup F$ is clearly an embedding of $P_0 \cup F$. The set $E_1=\{r_2,r_4,r_8,$ $r_{11},r_{12},$ $r_{13}, r_{14}\}$ is also an embedding of $grnd(P_0) \cup F$. Indeed, for every rule $r \in E_1$, $E_1 \embeds_h r$, and for every rule $r \in (grnd(P_0) \cup F)\setminus E_1$ it holds that $E_1\nembeds_b r$.
The set $E_2=\{r_4,r_8,r_{12},r_{13},r_{14} \}$ is not an embedding of $grnd(P_0) \cup F$; indeed, $E_2\nembeds r_2$ since $E_2\embeds_b r_2$ and $E_2\nembeds_h r_2$.

%The set $E_5=F_1\cup\{r(a,b),r(b,b)\}$ is the minimal embedding of $F_1\cup grnd(P_0)$ and it coincides with the unique answer set (as well as the perfect model) of $F_1\cup P_0$.

%
%Indeed $\foreach r\in F_1\cup grnd(P_0)\setminus E_2$, $E_2 \nembeds_b r$, thus $E_2\embeds r$. The set $E_3=F_1\cup\{r_2,r_3,r_4,r_8,r_{12}\}$ is also an embedding of $F_1\cup grnd(P_0)$ as similarly to $E_2$, for every rule $r \in \{r_2,r_4,r_8,r_{12}\}$ $E_3\embeds r$ and $E_3\nembeds r_3$.

\end{example}

\begin{table}[t]
        \begin{tabular}{l|l}
        \hline

        \multicolumn{2}{c}{Let $P$ be a program, $F$ a set of facts, and $r \in \grnd(P) \cup F$} \\

        \hline

        {\emph{\textbf{Model-Theoretic Semantics}}} & {\emph{\textbf{Embedding Programs Semantics}}}\\

        & \\

        $I :$ Set of {\em atoms} & $S : $ Set of {\em rules} \\

        %\footnotesize{$I \models B(r)$, if $B(r) \in I$}&\footnotesize{$S \embeds_b r$, if $\forall a \in B^+(r)$ $\exists r' \in S$ s.t. $a \in H(r')$}\\

        \footnotesize{$I \models B(r)$, if $B^+(r) \subseteq I$ and
                   $B^-(r) \cap \{\text{not}\ a\ |\ a \in I\} = \emptyset$}&\footnotesize{$S \embeds_b r$, if $\forall a \in B^+(r)$ $\exists r' \in S$ s.t. $a \in H(r')$}\\

        \footnotesize{$I \models H(r)$, if $H(r) \in I$}&\footnotesize{$S \embeds_h r$, if $r \in S$}\\
        \footnotesize{$I \models r$, if either: $(i)$ $I \nvDash B(r)$, or $(ii)$ $I \models H(r)$}&\footnotesize{$S \embeds r$, if either: $(i)$ $S \nembeds_b r$, or $(ii)$ $S \embeds_h r$}\\
        \hline
        \end{tabular}
        \caption{\label{fig:embcomparison} Comparison of the classic model-theoretic semantics for logic programs and the embedding program semantics.}
\end{table}

%It turns out that embedding programs enjoy some nice properties.
%\todo{GB: Esempio di embedding qui. Il lemma su $A \subseteq Heads(P)$ \`{e} stato spostato in appendice}
%\todo{Esempio di embedding aggiunto. Check}

Embedding programs enjoy a number of interesting properties, some of which are reported next. First, an embedding program is equivalent to
$P \cup F$ (Theorem~\ref{theo:Embequivalence}); also, an intersection of embedding programs is an embedding program, similarly to the intersection of models (Proposition~\ref{prop:IntersectionEP}).

\begin{theorem}\label{theo:Embequivalence}\rm (Embedding equivalence).
For a program $P$, a set of facts $F$ and an embedding program $\EP{}$ for $P \cup F$, $AS(\grnd(P)\cup F)$ = $AS(\EP{})$.
\end{theorem}

\begin{proposition}\label{prop:IntersectionEP} \rm (Intersection of embedding programs).
Given a logic program $P$, a set of facts $F$, $\EP{1}$ and $\EP{2}$ embedding programs for $P \cup F$,
$\EP{}= \EP{1} \cap \EP{2}$ is an embedding program for $P \cup F$.
\end{proposition}

\begin{example}\label{exmp:embedding2}
Consider again the program $P_0$, the set of facts $F=\{e(a,b),e(b,b)\}$ and the embedding $E_1$ of the example above. It is easy to see that the set of rules $E_3=\{r_2,r_4,r_7,r_8,r_{12},r_{13},r_{14}\}$ is also an embedding of $F\cup grnd(P_0)$, and that $E_4= E_1\cap E_3 = \{r_2,r_4,r_8,r_{12},r_{13},r_{14}\}$ is an embedding for $F\cup grnd(P_0)$ as well.
\end{example}

Finally, the next theoretical results show that $P \cup F$ has a minimal embedding program, corresponding to the intersection of all embedding programs. Also, the minimal embedding program can be computed as the fixpoint of $\inst$, thus establishing a correspondence between embedding programs and the operational semantics of grounders.
%It comes out that, modulo facts, any embedding program contains the application of $\inst$ to $P$ with respect to the embedding itself.

\begin{theorem}\label{theo:fixedpointsareEPs} \rm
Given a program $P$ and a set of facts $F$, $\EP{} \subseteq \grnd(P) \cup F$ is an embedding program for $P \cup F$ iff $\EP{} \supseteq \instt{P}{\EP{}}\cup F$.
\end{theorem}

%The next theorem follows.

\begin{theorem}\label{cor:intersectionOfEmbedding}\rm
Let $\EPS$ be the set of embeddings of $P\cup F$; then,\ \
\label{cor:lfpisanembedding}
\[
\insttinf{P}{F}\cup F = \bigcap_{\EP{} \in \EPS} \EP{}.
\]
\end{theorem}

\begin{example}\label{exmp:embedding3}
Note that $\insttinf{P_0}{F}\cup F$ for the program $P_0$ of example \ref{exmp:embedding2} coincides with
$E_4$. It can be verified that $E_4$ corresponds to the intersection of all embedding programs for $F\cup grnd(P_0)$. In order to grasp the intuition, one could note that rules $r_{13}$ and $r_{14}$ belong to all embeddings as they are facts; moreover, rule $r_2$, $r_4$, $r_8$ and $r_{12}$ must belong to all embeddings as well, as their bodies are embedded by any set of rules containing $F$.
\end{example}

%
% QUESTO TEOREMA IN FUTURO SI PUO' TOGLIERE ABBIAMO GLI ALTRI CHE LO SUSSUMONO
%
\begin{theorem}\label{theo:intersequivalence} \rm
Given a logic program $P$ and a set of facts $F$, let $\EPS$ be the set of embeddings of $P\cup F$. Then
\[ AS(P \cup F) = AS\big(\bigcap_{\EP{} \in \EPS} \EP{}\,\big). \]
\end{theorem}

Note that Theorem~\ref{cor:intersectionOfEmbedding} combined with Theorem~\ref{theo:Embequivalence} constitutes an alternative, cleaner proof of Theorem~\ref{theo:instInfCorretto}.

%\begin{theorem}\label{theo:allEmbequivalence} \rm
%Given a logic program $P$, a set of facts $F$, and an embedding $\EP{}$ of $P\cup F$. Then $ AS(P \cup F) = AS(\EP{})$.
%\end{theorem}
%\begin{proof}
%Let $IF = \insttinf{P}{F} \cup F$.
%We prove the theorem by showing that for an answer set $A \in AS(IF)$ or $A \in AS(\EP{})$ the two FLP reducts $IF^A$ and $\EP{}^A$ coincide.
%Clearly, $IF^A \subseteq \EP{}^A$ by Theorem~\ref{cor:intersectionOfEmbedding}.
%By contradiction, let $r \in \EP{}^A \setminus IF^A$: clearly, $r \not\in IF$.
%
%[Case 1:\ $A \in AS(IF)$].\ \ \ Since $IF$ is an embedding, then $IF \embeds r$ and in particular $IF \not\embeds_b r$. Thus, there exists $a \in B^+(r)$ s.t. $a \not\in Heads(IF)$. Clearly $a \not\in A$, hence $r$ cannot appear in $\EP{}^A$.
%
%[Case 2:\ $A \in AS(\EP{})$].\ \ \ Since $r \in \EP{}^A$ but $r \not\in IF$, there exists at least one atom $a \in B^+(r)$ such that $a \not\in Heads(IF)$.
%On the other hand, $a \in A$, and by Th. 6.22 of~\cite{DBLP:journals/iandc/LeoneRS97}, if must belong to the last step of a so-called computation (i.e., a well-founded sequence of monotonically increasing interpretations whose first element contains $F$).\todo{nota che le computation le abbiamo gia' citate per un'altra proof -- uniformiamo il modo in cui ne parliamo}
%However, $F \cup IF$, thus a contradiction arises.
%\end{proof}

%%%%%%%%%%%%%%%%%%%%%%%%%%%%%%%%%%%%%%%%%%%%%%%%%%%%%%%%%%%%%%%%%%%%%%%%%%%
%%%%%%%%%%%%%%%%%%%%%%%%%%%%%%%%%%%%%%%%%%%%%%%%%%%%%%%%%%%%%%%%%%%%%%%%%%%
\section{Overgrounding}
\label{sec:overgrounding}
%Let $S$ be a set of ground atoms or ground rules. Let $\Inst(S)$ be defined as

%\[ \Inst(\KB,S) = \{ r \in grnd(\KB) s.t. B^+(r) \subseteq S \}  \]

%whenever $S$ is intended as a set of rules, with slight abuse of notation
%we will implicitly define $\Inst_{\KB}(R)$ as equivalent to $\Inst_{\KB}(Heads(R))$.

%The above operator can be seen as a way for generating and selecting only ground rules that can be built by using a set of allowed ground atoms $S$. If $S$ is initially set to a set of input facts $F$, one can obtain a bottom-up constructed ground program equivalent to $\KB \cup F$ by iteratively applying $\Inst$.

%\begin{restatable}{theorem}{}[adapted from~\cite{DBLP:conf/iclp/CalimeriCIL08}]% \rm
%For a set of facts $F$
%, we define $\Inst^0(\KB,F)$ = $\Inst(\KB,F)$, and $\Inst^{k+1} = \Inst(\Inst^{k})$, for any natural $k$.
%The sequence $I^k$ converges in a finite number of steps to a finite fixed point $\Inst^\infty(\emptyset)$ and
%\[ AS(\Inst^\infty(\KB,F)) = AS(grnd(\KB \cup F)) \]
%\end{restatable}

In the following we assume we are given a program $P$ and a sequence of sets of facts $F_1$, ..., $F_n$; then, let us assume that we need to perform a series of distinct evaluations over a different $F_i$ at each shot.
In other words we aim at computing all the sets $AS(P \cup F_1)$, ..., $AS(P \cup F_n)$.

\begin{definition}
For an integer $k$, s.t. $1\leq k \leq n$, we define ${\UF}_k = \bigcup_{1 \leq i \leq k} F_i$ as the sets {\em accumulated facts} at shot $k$. Moreover, we define $G_k = \insttinf{P}{{\UF}_{k}}$ as the {\em overgrounded program} at shot $k$.
\end{definition}

Each overgrounded program $G_k$ is equivalent to $P \cup F_i$ for $1 \leq i \leq k$.

\begin{theorem}
\label{theo:core}
The following two statements hold:

%for each $k$,\
%\begin{equation}
%\label{eq1}
%\begin{compactenum}
%\item[$(1)$\ ]
$(1)$:\ \
$\insttinf{P}{{\UF}_{k-1}} \subseteq \insttinf{P}{{\UF}_{k}}$
%\end{equation}
%and {\em b)}
%\begin{equation}\label{eq2}
%    \[
; \hfill
%\item[$(2)$\ ]
$(2)$:\ \
$AS( \insttinf{P}{{\UF}_k} \cup F_i ) = AS(P \cup F_i )$.
%%\]
%\end{equation}
%\end{compactenum}
\end{theorem}

We can now devise an incremental grounding strategy relying on the theoretical results illustrated so far.
For the sake of simplicity, we illustrate the core of the idea and omit all technical details and optimizations about how to instantiate a program, referring the reader to the vast literature.
At each shot $k$, we keep the set of accumulated facts $\UF_k$ and the overgrounded program $G_k$ by incrementally updating $\UF_{k-1}$ and $G_{k-1}$.
In this setting, $AS(P \cup F_k)$ can be obtained by computing $AS(G_{k} \cup F_k)$.
More in detail, overgrounded programs are managed as follows:
%
% PER FAVORE MANTENERE QUESTA ITEMIZE IN UNA SOLA PAGINA
%
\begin{compactitem}
\item
    At shot $1$, we set $\UF_1 = F_1$, and $G_1 = \insttinf{P}{{\UF}_1}$
\item
    At generic shot $k$:
    \begin{compactenum}
    \item
        we set $\UF_k = \UF_{k-1} \cup F_k$,
    \item
        we compute a set of additional ground rules $\Delta G_k$, and
    \item
        we set $G_k = G_{k-i} \cup \Delta G_k$.
    \end{compactenum}
\end{compactitem}
%
% PER FAVORE MANTENERE QUESTA ITEMIZE IN UNA SOLA PAGINA
%

The computation of $\Delta G_k$ can be efficiently performed by using an optimized incremental algorithm that takes in input the newly added facts $F_k \setminus \UF_{k-1}$ and produces the new rules appearing in $G_k \setminus G_{k-1}$.
In our case, we developed a variant of the typical incremental iteration of the semi-naive algorithm~\cite{DBLP:books/cs/Ullman88}.

\newcommand\tab{\hskip2em}

\begin{figure*}[t]
\begin{center}
  \includegraphics[width=0.6\textwidth,keepaspectratio]{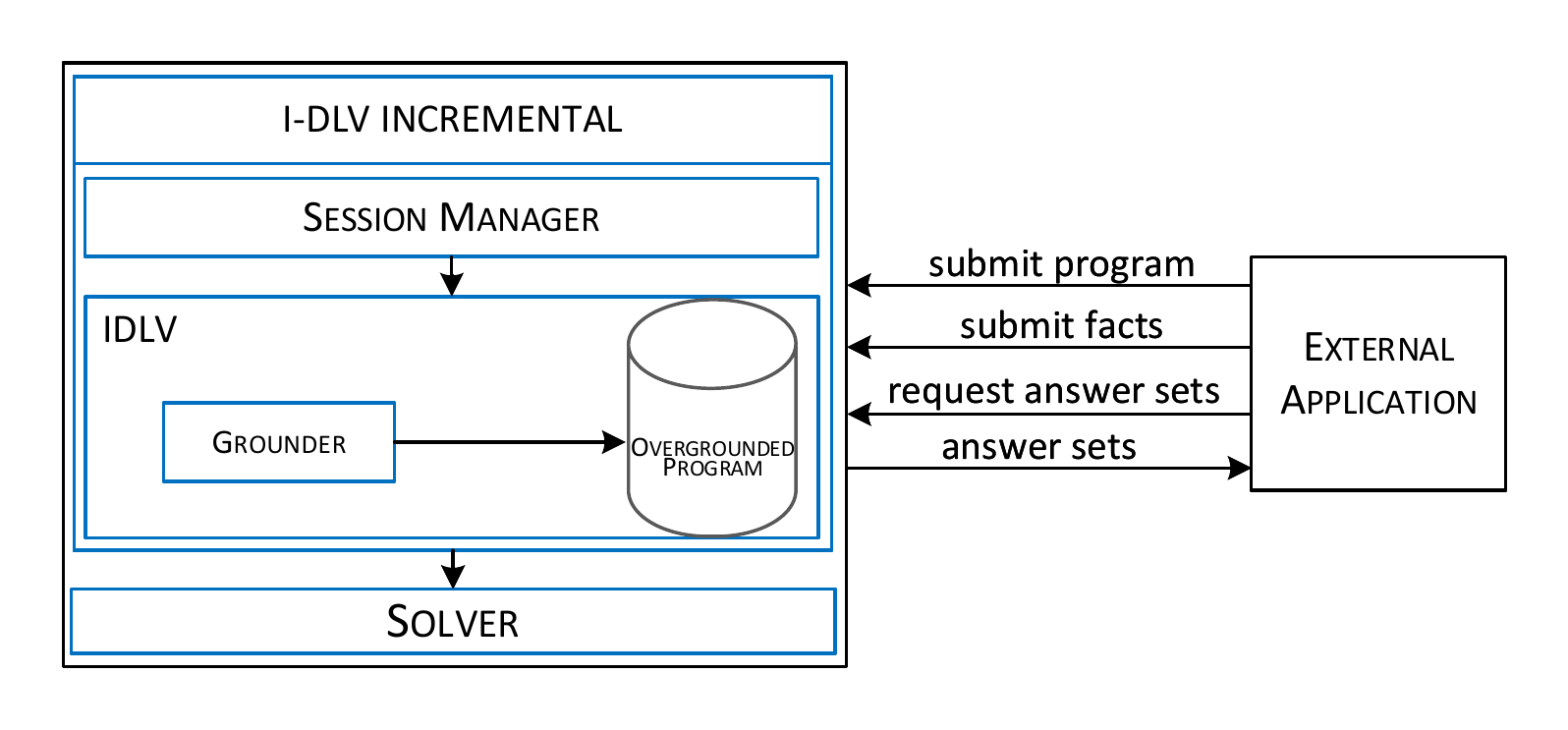}
  \caption{
 An infrastructure for incremental grounding.}
 \label{fig:Arch}
 \end{center}
\end{figure*}

\vspace*{-4.5em}

\section{Prototype structure and experimental evaluation}
The overgrounding strategy described above has been implemented by adapting the \idlv grounder~\cite{DBLP:journals/ia/CalimeriFPZ17,cali-perr-zang-TPLP-optimizing}; %to the purpose;
the architecture of the resulting prototype, called \system, is depicted in Figure~\ref{fig:Arch}.
%
%We implemented the overgrounding approach in a prototype called \system, whose architecture is depicted in Figure~\ref{fig:Arch}.
The system behaves as a process staying alive and providing a service-oriented behaviour, waiting for requests.
An external application $EA$ can open a working session and specify tasks to be carried out; working sessions are handled by a {\sc Session Manager} component.
After submitting a load request for a logic program $P$ along with an initial set of facts $F_1$, $EA$ can ask to compute the answer sets of $P$ over $F_1$: the {\sc Grounder} component is in charge of producing and storing the overgrounded program $\insttinf{P}{F_1}$; an external {\sc Solver} system is adopted to compute the answer sets of $\insttinf{P}{F_1} \cup F_1$.
This process can be repeated/iterated: $EA$ can provide additional sets of facts $F_k$ for $2\leq k\leq n$ so that $\insttinf{P}{\UF_k}$ with ${\UF}_k = \bigcup_{1 \leq i \leq k} F_i$ is incrementally produced and stored.
At each step $k$, the system is in charge of internally managing incremental grounding steps and automatically optimizing the computation by avoiding the re-instantiation of ground rules generated in a step $i<k$.
Again, the solver can be invoked to compute the answer sets of $\insttinf{P}{\UF_k} \cup F_k$.

The overgrounding approach peculiarly trades off memory for computation time spent during the grounding stage.
However, one might wonder about several questions: how big is the expected memory overhead in real scenarios, so to show in which contexts the increased memory demand can be considered acceptable; whether and how much grounding times decrease when overgrounding become increasingly larger, and thus rich of re-usable rules; whether and how much model generation times increase due to increasingly larger inputs.
Indeed, it is reasonable to expect performance worsening in the model generation phase, because of the larger number of non-simplified rules to process.

In order to assess the above issues, we conducted an experimental evaluation, considering two specific benchmarks taken from two real world settings~\cite{calimeri2013eternal,DBLP:conf/ruleml/CalimeriGIPPZ18} with different specific features: \pacman and Sudoku. The two benchmarks  constitute good and generalizable real cases of incremental scenarios:  the Sudoku domain allows to  perform a stress-test of the approach on grounding-intensive tasks; the Pacman game allows to assess effectiveness of overgrounding for real-time reasoning without the need for manual and involved customizations.
Experiments on Sudoku have been performed on a NUMA machine equipped with two $2.8$GHz AMD Opteron 6320 processors and 128GB of RAM, while experiments on \pacman have been performed on a machine equipped with a $2.2$GHz Intel Xeon Processor E5-2650 v4 and 64GB of RAM.

\pgfplotstableread[col sep=semicolon]{./test_results/sum-all-instances.csv}\datatable
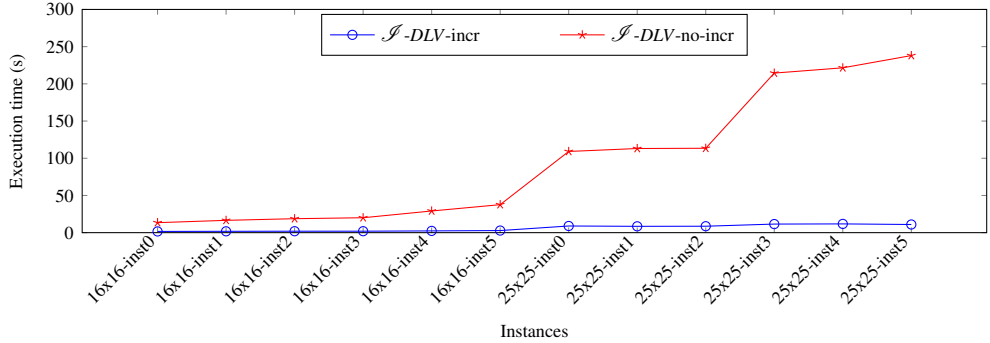
\begin{figure}[ht!]\centering
	\begin{tikzpicture}[scale=0.99]
	\pgfkeys{%
		/pgf/number format/set thousands separator = {}}
	\begin{axis}[
    	scale only axis
    	, font=\scriptsize
    	, x label style = {at={(axis description cs:0.5,-0.2)}}
    	, y label style = {at={(axis description cs:0.03,0.5)}}
    	, xlabel={Instances}
    	, ylabel={Execution time (s)}
    	, width=0.9\textwidth
        , legend style={at={(0.20,0.95)},anchor=north}
    	, height=3cm
    	, ymin=0, ymax=300
    	, ytick={0,50,100,150,200,250,300}
    	, major tick length=2pt
    	, xtick=data,
        , legend style={at={(0.5,0.98)},anchor=north, legend columns=2, /tikz/every even column/.append style={column sep=1.0cm}}
        , xticklabels from table={\datatable}{Instance}
        , x tick label style={rotate=45,anchor=east}
	]
	\addplot [mark size=1.75pt, color=blue, mark=o] [unbounded coords=jump] table[col sep=semicolon, x expr=\coordindex, y=Incremental] {./test_results/sum-all-instances.csv};
	\addlegendentry{\system}	
	\addplot [mark size=1.75pt, color=red, mark=star] [unbounded coords=jump] table[col sep=semicolon, x expr=\coordindex, y=No-Incremental] {./test_results/sum-all-instances.csv};
	\addlegendentry{\systemnoincr}
\end{axis}
	\end{tikzpicture}
\caption{
 Experiments on Sudoku benchmarks.\label{fig:exp1}}
\end{figure}

\vspace{-6em}

\subsection{Multi-shot Sudoku}
The classic Sudoku puzzle, or simply ``Sudoku'', consists of a tableau featuring $81$ cells, or positions, arranged in a $9$ by $9$ grid.
%The grid is divided into nine sub-tableaux (regions, or blocks) containing nine positions each. Initially, in the game setup a number of positions are filled with a number between $1$ and $9$. The problem consists in checking whether the empty positions can be filled with numbers in a way such that each row, each column and each block shows all digits from $1$ to $9$ exactly once.
When solving a Sudoku, players typically adopt deterministic inference strategies allowing, possibly, to obtain a solution. Several deterministic strategies are known~\cite{calimeri2013eternal} and can be encoded in ASP; herein, we took into account two simple strategies, namely, ``naked single'' and ``hidden single''.
%The former one permits to entail that a number $n$ has to be associated to a cell $C$ when all other numbers are excluded to be in $C$; for instance, in a Sudoku of $9$ rows and $9$ columns, assuming that we inferred that all numbers between $1$ and $8$ cannot be in the cell $(1,1)$, then, it must contain $9$.
%The hidden single strategy, instead, allows to derive that only a cell of a row/column/block can be associated with a particular number; for instance, in a Sudoku of $9$ rows and $9$ columns, the only cell that can contain $3$ is $(4,5)$ if, according to Sudoku rules, all other cells in the same block of $(4,5)$, row $4$ and column $5$ cannot hold the number $3$.

The encoding of Sudoku deterministic inference rules is generally grounding-intensive, like in the following code fragment, encoding the naked single strategy:
%\footnote{Complete encodings are available at~\url{https://github.com/DeMaCS-UNICAL/Incremental-answer-set-programming-with-overgrounding/wiki}.}:
%~\cite{incrwebsite}.}:
%\begin{scriptsize}
%\begin{verbatim}
\begin{lstlisting}[style=asp-style]
candidatesAreMoreThan2(X,Y):-candidate(X,Y,N),candidate(X,Y,N1),N!=N1.
newValue(X,Y,N):-candidate(X,Y,N),not candidatesAreMoreThan2(X,Y),nogiven(X,Y).
\end{lstlisting}
%\end{verbatim}
%\end{scriptsize}
where an atom {\em candidate(X,Y,N)} specifies that number $N$ can be possibly assigned to the cell $(X,Y)$, and {\em nogiven(X,Y)} tells that the value of the cell $(X,Y)$ has not been inferred yet.

%The iterated application of inference rules to given Sudoku tables is a good test for appreciating the impact of the incremental evaluation, since updated Sudoku tables contain all logical assertions derived in previous iterations.

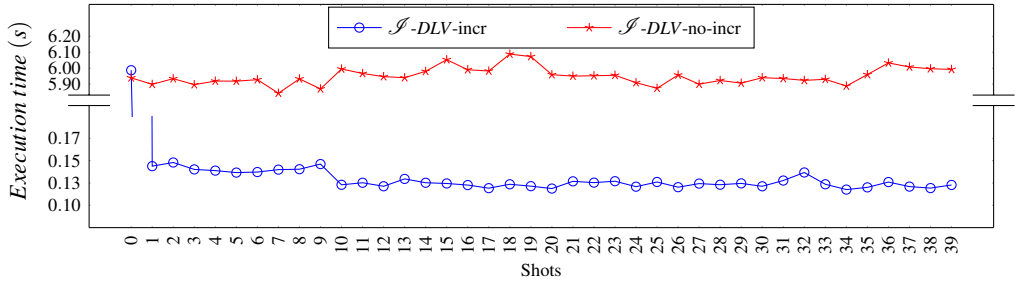
\begin{figure}[ht!]
\begin{tikzpicture}[scale=0.99]
\pgfkeys{%
        /pgf/number format/set thousands separator = {}}
\begin{groupplot}[
    group style={
        group name=my fancy plots,
        group size=1 by 2,
        xticklabels at=edge bottom,
        vertical sep=0pt,
    }
    , xtick={0,1,...,39}
    , x label style = {at={(axis description cs:0.5,0.1)}}
    , x tick label style={rotate=90,anchor=east}
    , xmin=-2, xmax=41
    , width=0.9\textwidth
    , height=1.5cm
    , y label style = {at={(axis description cs:0.03,0.5)}}
    , legend style={at={(0.5,0.98)},anchor=north, legend columns=2, /tikz/every even column/.append style={column sep=1.0cm}}
    , major tick length=0.5pt
]

\nextgroupplot[scale only axis,
               ymin=5.7,ymax=6.4,
               ytick={5.9,6.0,...,6.3},
               font=\scriptsize,
               y tick label style={/pgf/number format/.cd, fixed, fixed zerofill, precision=2, /tikz/.cd},
               axis x line*=bottom,axis x line*=top,
               axis y discontinuity=parallel
              ]
	\addplot [mark size=1.75pt, color=blue, mark=o] [unbounded coords=jump] table[col sep=semicolon, x expr=\coordindex, y=Incremental] {./test_results/all-iterations-instance5_25x25.csv};
	\addlegendentry{\system}
	\addplot [mark size=1.75pt, color=red, mark=star] [unbounded coords=jump] table[col sep=semicolon, x expr=\coordindex, y=No-Incremental] {./test_results/all-iterations-instance5_25x25.csv};
	\addlegendentry{\systemnoincr}
\coordinate (top) at (rel axis cs:0,1);

    \nextgroupplot[scale only axis,
               ymin=0.075,ymax=0.2,
               ytick={0.100,0.125,...,0.175},
               font=\scriptsize,
               axis x line*=top,axis x line*=none,
               xlabel={Shots},
               y tick label style={/pgf/number format/.cd, fixed, fixed zerofill, precision=2, /tikz/.cd}
               ]
	\addplot [mark size=1.75pt, color=blue, mark=o] [unbounded coords=jump] table[col sep=semicolon, x expr=\coordindex, y=Incremental] {./test_results/all-iterations-instance5_25x25.csv};
	\addplot [mark size=1.75pt, color=red, mark=star] [unbounded coords=jump] table[col sep=semicolon, x expr=\coordindex, y=No-Incremental] {./test_results/all-iterations-instance5_25x25.csv};
    \coordinate (bot) at (rel axis cs:1,0);

\end{groupplot}

\path (top-|current bounding box.west)--
            node[anchor=south,rotate=90] {\(Execution\ time\ (s)\)}
            (bot-|current bounding box.west);

\end{tikzpicture}
\vspace{-32em}
\caption{Grounding times for all iterations of a 25x25 Sudoku instance.\label{fig:exp2}}
\end{figure}

In the experiments we considered generalized Sudoku tables of size $16$x$16$ and $25$x$25$, and tested logic programs under answer set semantics encoding deterministic inference rules.
We compared two different evaluation strategies: $(i)$ \system implementing the incremental approach, and $(ii)$ \systemnoincr in which no incremental evaluation policy is applied.
Both strategies have been executed in an online setting, in which consecutive series of input facts are submitted.
For a given Sudoku table, the two inference rules above were modelled via ASP logic programs (see ~\cite{calimeri2013eternal}).
The resulting answer set encodes a new tableau, possibly deriving new numbers to be associated to initially empty cells, and reflecting the application of inference rules; the new tableau is then given as input to the system and, again, by means of the same inferences, new cell values are possibly entailed.
The process is iterated until no further association is found.
In general, given a Sudoku, it cannot be assumed that the deterministic approach leads to a complete solution; however, for each considered Sudoku size, we selected only instances which are completely solvable with the two inference rules described above.

%\todo{ordinare per tempi crescenti, non per label dell'istanza}
Grounding times of \system and \systemnoincr are plotted in Figure~\ref{fig:exp1}: for each instance, the total grounding time (in seconds) computed over all iterations is reported.
\system required at most $12$ seconds to iteratively solve each instance, and performed clearly better than \systemnoincr that required up to $237$ seconds, instead; this results in an improvement of $95\%$. %\todo{rivedere tempi con risultati finali. Chiarire se solve significa fare anche model generation}
Figure~\ref{fig:exp2} shows a closer look on the performance obtained in the instance $4$ of size $25$x$25$ which is the one requiring the highest amount of time to be solved and the highest number of iterations: for each iteration, the grounding time (in seconds) is reported.
At the first iteration, both configurations spent almost the same time; for each further iteration, though, \system required an average time of $0.13$ seconds: a time reduction of 98\% w.r.t. \systemnoincr, that showed an average time of about $5.95$ seconds. On the overall, this confirms the potential of our incremental grounding approach in scenarios involving updates in the underlying input facts.

\subsection{Multi-shot Pac-Man}
In our work~\cite{DBLP:conf/ruleml/CalimeriGIPPZ18} we show how to integrate an ASP-based reasoning module in the Unity game development framework, and showcase an artificial player for the classic real-time game \pacman. %, whose intelligence is encoded using a logic program $P_{pac}$.
The \pacman moves in a board containing a number of pellets and four ghosts chasing him: the goal is to eat all pellets while avoiding the four ghosts.
The \pacman must continuously decide the way to take, depending on ``dynamic'' (i.e., changing at each reasoning shot) facts representing the current status of the board (empty/pellet tiles, positions of the four ghosts).
The new direction of the \pacman depends on which logical assertion among {\em next(up)}, {\em next(down)}, {\em next(left)}, {\em next(right)} belongs in the \quo{best} guessed answer set.
%
%
%$P_{pac}$ takes as ``dynamic'' (i.e., changing at every reasoning input facts the current status of the board (empty/pellet tiles, positions of the four ghosts). The new direction of the \pacman depends on which of the four logical assertions {\em next(up)}, {\em next(down)}, {\em next(left)}, {\em next(right)} belongs in the ``best'' answer set.
%
%The \pacman must continuously decide which direction to take in the game board, so
%The
%$P_{pac}$ is subject to
%
The intelligence of the \pacman is encoded via a logic program $P_{pac}$, that must be repeatedly executed; this requires multiple grounding+solving jobs over slightly different and unforeseen inputs.

%We benchmarked $P_{pac}$ with our new overgrounding engine.
%Hence, our solution consists in executing the reasoner module only after that the proposed move has been completed on the graphic side.

%In the following, we briefly describe the aforementioned reasoning module; it is worth remarking how this work is not intended at the development of a state-of-the-art artificial \pacman player, but rather at showing the viability of our approach.

%\medskip
For space reasons,
%we refrain from going into a complete description of $P_{pac}$; rather,
we focus next only on parts of $P_{pac}$ requiring a significant effort at the grounding stage.
%\footnote{The complete logic programs are available at~\cite{incrwebsite}.}.
%However, it is worth noting how the declarative approach allows to easily and quickly define an AI and facilitates changes in the strategy by simply modifying some rules.
%For instance, weak constraints in the optimizing part may guide the behaviour of the \pacman player by considering different factors, along with their importance.
%Hence, just by focusing on the optimizing part, by means of just a few changes we may significantly change the behaviour of the \pacman player, as shown next.
%
Instances of the {\em tile} predicate encode the game board divided into tiles; the fact {\em pacman(x,y)} represents the current position of the \pacman, while {\em ghost(x,y,g)} represents the position of ghost $g$; atoms of the form {\em nextTile(X,Y)} encode the possible next positions of the \pacman.
The strategy adopted by $P_{pac}$ is quite simple: priority is to get away from ghosts.
To this end, distances between the next position candidates and all ghosts are computed: the next position chosen is among the ones increasing the distance to the nearest ghost.
%According to the strategy adopted in $P_{pac}$, the \pacman tries to get away from ghosts; to this end, it is needed to compute the distance between the position on which the \pacman is moving in the next step and each ghost: the next \pacman direction is picked among those increasing the distance between the \pacman and the nearest ghost.
This behaviour is achieved via an ASP code fragment similar to the following:
%
%\begin{scriptsize}
%\begin{verbatim}
\begin{lstlisting}[style=asp-style]
nextTile(X,Y) :- pacman(Px,Y), next(right), X=Px+1, tile(X,Y).
nextTile(X,Y) :- pacman(Px,Y), next(left), X=Px-1, tile(X,Y).
nextTile(X,Y) :- pacman(X,Py), next(up), Y=Py+1, tile(X,Y).
nextTile(X,Y) :- pacman(X,Py), next(down), Y=Py-1, tile(X,Y).
adjacent(X1,Y1,X2,Y2) :- tile(X1,Y1), tile(X2,Y2), step(DX,DY), X2=X1+DX, Y2=Y1+DY.
adjacent(X1,Y1,X2,Y2) :- tile(X1,Y1), tile(X2,Y2), step(DX,DY), X2=X1-DX, Y2=Y1-DY.
distance(X1,Y1,X2,Y2,1) :- tile(X1,Y1), adjacent(X1,Y1,X2,Y2).
distance(X1,Y1,X3,Y3,D) :- number(D), distance(X1,Y1,X2,Y2,D-1), adjacent(X2,Y2,X3,Y3).
distPacmanGhost(D,G) :- nextTile(Xp,Yp), ghost(Xg,Yg,G), minDistance(Xp,Yp,Xg,Yg,D).
noMinDistPacmanGhost(X) :- distPacmanGhost(X,_), distPacmanGhost(Y,_), distance(X), X>Y.
minDistancePacmanNextGhost(MD) :- not noMinDistPacmanGhost(MD), distPacmanGhost(MD,_).
\end{lstlisting}
%\end{verbatim}
%\end{scriptsize}
%Lines $3-4$ compute pairs of adjacent cells of radius $1$. Lines $5-6$ determine distances among tiles: given that the predicate \texttt{number} ranges from $1$ to $10$, for each tile, we are computing all possible distances with tiles in a radius of at most $10$.
%
The above code contains parts which will likely have the same instantiation regardless of input facts (e.g., the {\em adjacent} and {\em distance} predicates), and parts whose instantiation will slightly differ depending on input facts (i.e., {\em nextTile} and {\em distPacmanGhost}, which depend on current positions of \pacman and ghosts).
Such two categories of code fragments would roughly correspond to parts respectively marked with the former {\tt \#base} and {\tt \#cumulative} directives of iclingo~\cite{DBLP:journals/tplp/GebserKKS19,DBLP:conf/lpnmr/GebserKKS11}, and later generalized with the {\tt \#program} keyword of clingo; however, it is not necessary to introduce specific grounding directives within ASP code while using \system.
In~\cite{DBLP:conf/ruleml/CalimeriGIPPZ18}, in the absence of an overgrounding engine, we manually tabled the instantiation of the logic program above, we resorted to procedural code for many aspects of the reasoning process, and we limited the maximum visibility horizon of the \pacman to $10$ tiles.
%This allowed us to achieve a reasonable performance.
By adopting the overgrounding approach, we were able to avoid manual optimizations and achieved a fully automatic incremental approach: the grounded program is internally stored right after the first computation, thus bypassing re-computations of heavy grounding tasks.
%Intuitively, alternative definitions, improving grounding times, would be in principle possible\footnote{For instance, an atom of the form {\em nextTile(X1,Y1)} could be added in the bodies of rules defining {\em distance}, thus reducing the number of instantiated rules.}, but at the price of loosing generality and requiring expert knowledge on the evaluation process of logic programs.
%In this context, one of the advantages of our overgrounding approach is to limit drawbacks stemming from non-optimized encodings, thus permitting the definition of complex sets of rules without the burden of tweaking and optimizing logic programs.

\def\scalatempi{0.84}
\begin{figure}[H]
\addtocounter{subfigure}{-1}
\sidesubfloat{
\begin{tikzpicture}[scale=0.99]
\pgfkeys{%
        /pgf/number format/set thousands separator = {}}
\begin{groupplot}[
    group style={
        group name=my fancy plots,
        group size=1 by 2,
        xticklabels at=edge bottom,
        vertical sep=0pt,
    }
    , xtick={0,20,...,460}
    , xmin=-2, xmax=470
    , width=\scalatempi\textwidth
    , x label style = {at={(axis description cs:0.5,0.1)}}
    , height=1.5cm
    , y label style = {at={(axis description cs:0.03,0.5)}}
    , legend style={at={(0.5,0.98)},anchor=north, legend columns=2, /tikz/every even column/.append style={column sep=1.0cm}}
    , major tick length=0.5pt
]

\nextgroupplot[scale only axis,
               ymin=64,ymax=73,
               ytick={66,68,...,72},
               font=\scriptsize,
               height=2cm,
               y tick label style={/pgf/number format/.cd, fixed, fixed zerofill, precision=1, /tikz/.cd},
               axis x line*=bottom,axis x line*=top,
               axis y discontinuity=parallel
              ]
    \addplot [mark size=0.2pt, color=blue, mark=o] [unbounded coords=jump] table[col sep=semicolon, x expr=\coordindex, y=IDLV] {./benchmark/csv/non-incremental-30.csv};
    \addlegendentry{\systemnoincr}
    \addplot [mark size=0.2pt, color=teal, mark=o] [unbounded coords=jump] table[col sep=semicolon, x expr=\coordindex, y=IDLV] {./benchmark/csv/incremental-30.csv};
    \addlegendentry{\system}
\coordinate (top) at (rel axis cs:0,1);

    \nextgroupplot[scale only axis,
               ymin=2,ymax=11,
               ytick={4,6,...,10},
               font=\scriptsize,
               height=1cm,
               axis x line*=top,axis x line*=none,
               y tick label style={/pgf/number format/.cd, fixed, fixed zerofill, precision=1, /tikz/.cd}
               ]
    \addplot [mark size=0.2pt, color=blue, mark=o] [unbounded coords=jump] table[col sep=semicolon, x expr=\coordindex, y=IDLV] {./benchmark/csv/non-incremental-30.csv};
    \addplot [mark size=0.2pt, color=teal, mark=o] [unbounded coords=jump] table[col sep=semicolon, x expr=\coordindex, y=IDLV] {./benchmark/csv/incremental-30.csv};

\coordinate (bot) at (rel axis cs:1,0);

\end{groupplot}

\path (top-|current bounding box.west)--
            node[anchor=south] {\small\(\ \ \ \ \ \ \ \ \ \ \)}
            (bot-|current bounding box.west);

\path (top-|current bounding box.west)--
            node[anchor=south] {\(\footnotesize\textbf{(a)\ \ \ \ \ \ }\)}
            (bot-|current bounding box.west);

\end{tikzpicture}}
\vspace{-33.5em}
\addtocounter{subfigure}{-1}
\sidesubfloat{
\begin{tikzpicture}[scale=0.99]
\pgfkeys{%
        /pgf/number format/set thousands separator = {}}
\begin{groupplot}[
    group style={
        group name=my fancy plots,
        group size=1 by 2,
        xticklabels at=edge bottom,
        vertical sep=0pt,
    }
    , xtick={0,20,...,460}
    , x label style = {at={(axis description cs:0.5,0.1)}}
    , xmin=-2, xmax=470
    , width=\scalatempi\textwidth
    , height=1.5cm
    , y label style = {at={(axis description cs:0.03,0.5)}}
    , legend style={at={(0.5,0.98)},anchor=north, legend columns=2, /tikz/every even column/.append style={column sep=1.0cm}}
    , major tick length=0.5pt
]

\nextgroupplot[scale only axis,
               ymin=1.75,ymax=7,
               ytick={3,3.75,...,6},
               font=\scriptsize,
               height=2cm,
               y tick label style={/pgf/number format/.cd, fixed, fixed zerofill, precision=2, /tikz/.cd},
               axis x line*=bottom,axis x line*=top,
               axis y discontinuity=parallel
              ]
    \addplot [mark size=0.2pt, color=green, mark=triangle*] [unbounded coords=jump] table[col sep=semicolon, x expr=\coordindex, y=WASP] {./benchmark/csv/non-incremental-30.csv};
    \addlegendentry{Wasp + classic grounding}
    \addplot [mark size=0.2pt, color=red, mark=star] [unbounded coords=jump] table[col sep=semicolon, x expr=\coordindex, y=CLASP] {./benchmark/csv/non-incremental-30.csv};
    \addlegendentry{Clasp + classic grounding}
    \addplot [mark size=0.2pt, color=brown, mark=triangle*] [unbounded coords=jump] table[col sep=semicolon, x expr=\coordindex, y=WASP] {./benchmark/csv/incremental-30.csv};
    \addlegendentry{Wasp + overgrounded program}
    \addplot [mark size=0.2pt, color=purple, mark=star] [unbounded coords=jump] table[col sep=semicolon, x expr=\coordindex, y=CLASP] {./benchmark/csv/incremental-30.csv};
    \addlegendentry{Clasp + overgrounded program}
\coordinate (top) at (rel axis cs:0,1);

    \nextgroupplot[scale only axis,
               ymin=0.15, ymax=1.1,
               ytick={0.25,0.50,...,1.0},
               font=\scriptsize,
               height=1.0cm,
               axis x line*=top,axis x line*=none,
               y tick label style={/pgf/number format/.cd, fixed, fixed zerofill, precision=2, /tikz/.cd}
               ]
    \addplot [mark size=0.2pt, color=green, mark=triangle*] [unbounded coords=jump] table[col sep=semicolon, x expr=\coordindex, y=WASP] {./benchmark/csv/non-incremental-30.csv};
    \addplot [mark size=0.2pt, color=red, mark=star] [unbounded coords=jump] table[col sep=semicolon, x expr=\coordindex, y=CLASP] {./benchmark/csv/non-incremental-30.csv};

\coordinate (bot) at (rel axis cs:1,0);

\end{groupplot}

\path (top-|current bounding box.west)--
            node[anchor=south,rotate=90] {\footnotesize\(Execution\ time\ (s)\)}
            (bot-|current bounding box.west);

\path (top-|current bounding box.west)--
            node[anchor=south] {\(\footnotesize\textbf{(b)\ \ \ \ \ \ }\)}
            (bot-|current bounding box.west);

\end{tikzpicture}}
\vspace{1em}
\addtocounter{subfigure}{-1}
\sidesubfloat{
\begin{tikzpicture}[scale=0.99]
\pgfkeys{%
        /pgf/number format/set thousands separator = {}}
\begin{groupplot}[
    group style={
        group name=my fancy plots,
        group size=1 by 2,
        xticklabels at=edge bottom,
        vertical sep=0pt,
    }
    , xtick={0,20,...,460}
    , x label style = {at={(axis description cs:0.5,0.1)}}
    , xmin=-2, xmax=470
    , width=\scalatempi\textwidth
    , height=1.5cm
    , y label style = {at={(axis description cs:0.03,0.5)}}
    , legend style={at={(0.5,0.98)},anchor=north, legend columns=2, /tikz/every even column/.append style={column sep=1.0cm}}
    , major tick length=0.5pt
]

\nextgroupplot[scale only axis,
               ymin=64,ymax=76,
               ytick={67,69,...,73},
               font=\scriptsize,
               height=2.5cm,
               y tick label style={/pgf/number format/.cd, fixed, fixed zerofill, precision=1, /tikz/.cd},
               axis x line*=bottom,axis x line*=top,
               axis y discontinuity=parallel
              ]
    \addplot [mark size=0.2pt, color=red, mark=square*] [unbounded coords=jump] table[col sep=semicolon, x expr=\coordindex, y=IDLV+WASP] {./benchmark/csv/non-incremental-30.csv};
    \addlegendentry{\systemnoincr + Wasp}
    \addplot [mark size=0.2pt, color=brown, mark=diamond*] [unbounded coords=jump] table[col sep=semicolon, x expr=\coordindex, y=IDLV+CLASP] {./benchmark/csv/non-incremental-30.csv};
    \addlegendentry{\systemnoincr + Clasp}
    \addplot [mark size=0.2pt, color=magenta, mark=square*] [unbounded coords=jump] table[col sep=semicolon, x expr=\coordindex, y=IDLV+WASP] {./benchmark/csv/incremental-30.csv};
    \addlegendentry{\system + Wasp}
    \addplot [mark size=0.2pt, color=blue, mark=diamond*] [unbounded coords=jump] table[col sep=semicolon, x expr=\coordindex, y=IDLV+CLASP] {./benchmark/csv/incremental-30.csv};
    \addlegendentry{\system + Clasp}
\coordinate (top) at (rel axis cs:0,1);

    \nextgroupplot[scale only axis,
               ymin=4.5,ymax=9.5,
               ytick={5,7,...,9},
               font=\scriptsize,
               height=0.5cm,
               xlabel={Shots},
               axis x line*=top,axis x line*=none,
               y tick label style={/pgf/number format/.cd, fixed, fixed zerofill, precision=1, /tikz/.cd}
               ]
    \addplot [mark size=0.2pt, color=red, mark=square*] [unbounded coords=jump] table[col sep=semicolon, x expr=\coordindex, y=IDLV+WASP] {./benchmark/csv/non-incremental-30.csv};
    \addplot [mark size=0.2pt, color=brown, mark=diamond*] [unbounded coords=jump] table[col sep=semicolon, x expr=\coordindex, y=IDLV+CLASP] {./benchmark/csv/non-incremental-30.csv};
    \addplot [mark size=0.2pt, color=magenta, mark=square*] [unbounded coords=jump] table[col sep=semicolon, x expr=\coordindex, y=IDLV+WASP] {./benchmark/csv/incremental-30.csv};
    \addplot [mark size=0.2pt, color=blue, mark=diamond*] [unbounded coords=jump] table[col sep=semicolon, x expr=\coordindex, y=IDLV+CLASP] {./benchmark/csv/incremental-30.csv};
\coordinate (bot) at (rel axis cs:1,0);

\end{groupplot}

\path (top-|current bounding box.west)--
            node[anchor=south] {\small\(\ \ \ \ \ \ \ \ \ \ \)}
            (bot-|current bounding box.west);

\path (top-|current bounding box.west)--
            node[anchor=south] {\(\footnotesize\textbf{(c)\ \ \ \ \ \ }\)}
            (bot-|current bounding box.west);

\end{tikzpicture}}
\vspace{-30em}
\caption{\label{fig:tempipacman}
\pacman benchmark: \textbf{(a)} Grounding time. \textbf{(b)} Solving Time. \textbf{(c)} Total time.}
\end{figure}
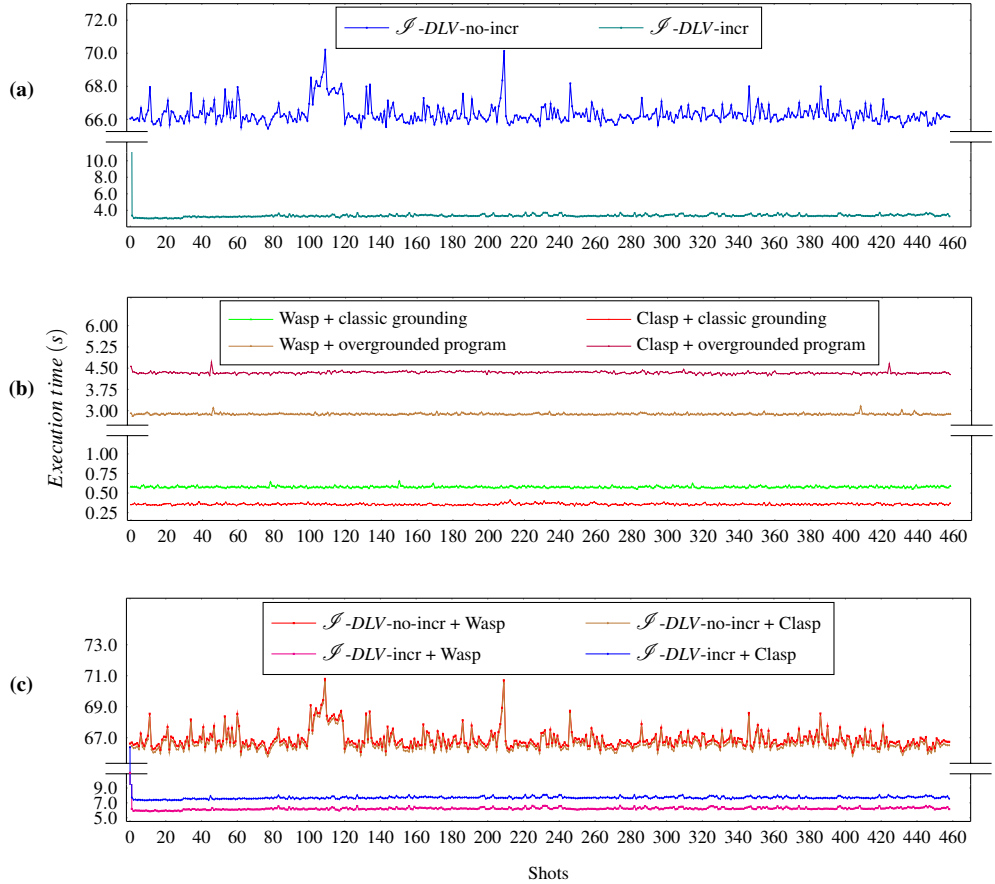

It is worth noting that the {\em adjacent} and {\em distance} predicates are defined in a general but inefficient way; this can likely be the case if such a predicate was taken from a predicate module library.
Alternative definitions, improving grounding times, would be in principle possible: for instance, the distance between tiles is encoded with the predicate $distance(X_1,Y_1, X_2, Y_2,D)$, where $D$ is computed for all couple of points $(X_1,Y_1)\times(X_2,Y_2)$;
this is what one can expect from a modeller not knowing, and probably not wishing to know how to optimize this code.
Pushing $nextTile$ atoms within rule bodies would allow to reduce the grounding size by limiting grounding only to actually necessary distance values. Nonetheless, besides making code less readable and less declarative, this modification would disrupt modularity, and it requires some expertise on operational details of grounders.
%Think, e.g., at having a general re-usable definition of the $distance$ predicate. This would need the expertise to add manual patches in order to make the instantiated program smaller.

Experiments in the \pacman domain were conducted by logging a series of $459$ consecutive steps taken during an actual game; each step encodes a different status of the game board in terms of logical assertions.
Such inputs were, in turn, run along with $P_{pac}$ in a controlled environment outside the game engine, averaging times over five separate runs.
Grounding was performed by our new overgrounding engine \system and by the non-incremental version \systemnoincr.
Even though \systemnoincr re-computes ground programs per each shot, instantiations are generally smaller and better tailored to the current shot; instead, \system re-uses overgrounded programs, which are generally larger.

The solving task was instead performed using WASP~\cite{DBLP:conf/lpnmr/AlvianoDLR15} and clasp~\cite{DBLP:conf/lpnmr/GebserKK0S15}.
In order to assess performance in the worst case scenario, we allowed the \pacman to have a visibility horizon of $30$ tiles in each direction.

\begin{figure}[ht!]
\centering
\pgfplotstableread[col sep=semicolon]{./benchmark/csv/incremental-30.csv}\datatable
\sidesubfloat[]{
	\begin{tikzpicture}[scale=0.99]
	\pgfkeys{%
		/pgf/number format/set thousands separator = {}}
	\begin{axis}[
    	scale only axis
    	, font=\footnotesize
    	, x label style = {at={(axis description cs:0.5,0.05)}}
        , y label style = {at={(axis description cs:-0.02,0.5)}}
    	, xlabel={Shots}
    	, ylabel={\(Execution\ time\ (s)\)}
    	, width=0.84\textwidth
        , legend style={at={(0.2,0.95)},anchor=north}
    	, height=3cm
    	, ymin=1209574, ymax=1210200
        , y tick label style={/pgf/number format/.cd, fixed, fixed zerofill, precision=6, /tikz/.cd}
    	, ytick={1209594,1209674,1209754,1209834,1209914,1209994,1210074,1210154,1210234}
    	, major tick length=2pt
    	, xtick={0,20,...,460}
        , xmin=-2, xmax=470
	]

\addplot [mark size=0.2pt, color=red, mark=start*] [unbounded coords=jump] table[col sep=semicolon, x expr=\coordindex, y=NumberOfRules] {./benchmark/csv/incremental-30.csv};
    \addlegendentry{\system}

\end{axis}
	\end{tikzpicture}} \quad

\sidesubfloat[]{
    \begin{tikzpicture}[scale=0.99, baseline]
    \pgfkeys{%
        /pgf/number format/set thousands separator = {}}
    \begin{axis}[
        scale only axis
        , font=\footnotesize
        , x label style = {at={(axis description cs:0.5,-0.1)}}
        , y label style = {at={(axis description cs:0.03,0.5)}}
        , ylabel={\(Memory\ peaks\ (mb)\)}
        , width=0.882125\textwidth
        , legend style={at={(0.5,0.99)},anchor=north, legend columns=2, /tikz/every even column/.append style={column sep=1.0cm}}
        , height=3cm
        , ymin=0, ymax=360
        , ytick={0,30,...,360}
        , major tick length=0.5pt
        , xtick=data,
        , xticklabels={10,20,30}
        , x label style = {at={(axis description cs:0.5,0.05)}}
        , xlabel={Size}
%        , xticklabels from table={\datatable}{SIZE}
    ]
    \addplot [mark size=1.75pt, color=red, mark=star] [unbounded coords=jump] table[col sep=semicolon, x expr=\coordindex, y=INCREMENTAL] {./benchmark/csv/memory-peaks.csv};
    \addlegendentry{\system}
    \addplot [mark size=1.75pt, color=blue, mark=o] [unbounded coords=jump] table[col sep=semicolon, x expr=\coordindex, y=NO INCREMENTAL] {./benchmark/csv/memory-peaks.csv};
    \addlegendentry{\systemnoincr}
\end{axis}
    \end{tikzpicture}}
\caption{
\label{fig:memoriapacman}
 \pacman benchmark: \textbf{(a)} total number of rules after each iteration. \textbf{(b)} Memory peaks.}
\end{figure}
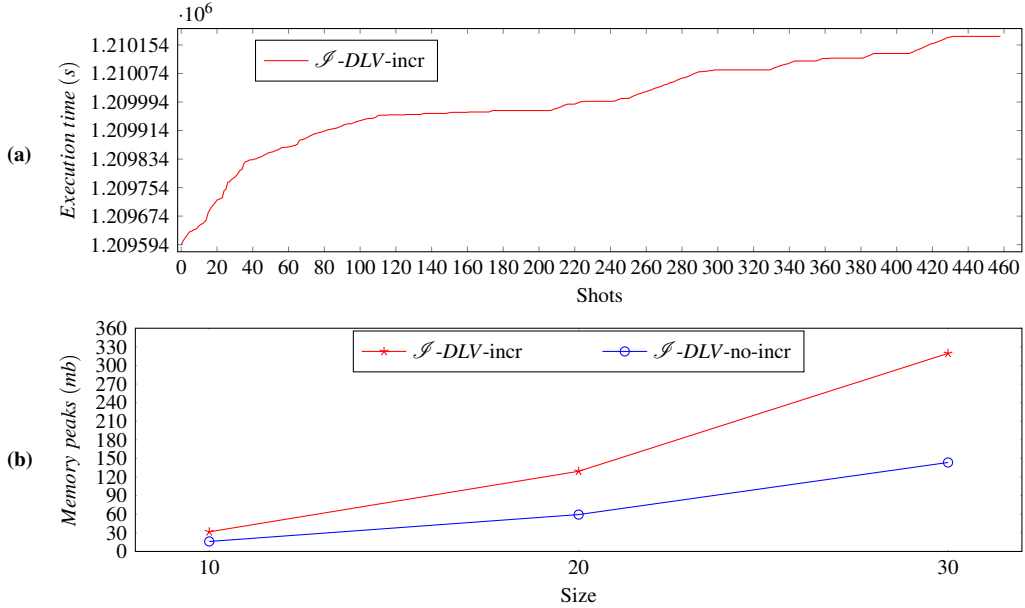

Results are depicted in Figure~\ref{fig:tempipacman} and Figure~\ref{fig:memoriapacman}; both show results on the $X$ axis in temporal execution order.
%Given the negligible fluctuations, we recorded $10$ shots and then the average baseline for experiments executed with \systemnoincr or instantiated programs thereof.
Figure~\ref{fig:tempipacman}~{\em (a)} compares instantiation times of both grounders, while
figure~\ref{fig:tempipacman}~{\em (b)} reports solver execution times when either an overgrounded program or a simplified instantiation is fed in input.
Both figures show that the overgrounding engine has similar performance to its non-incremental counterpart on the first iteration, while grounding times are considerably smaller for all subsequent iterations.
On the other hand, an acceptable worsening in solving times can be observed when overgrounded programs are fed in input.
Figure~\ref{fig:tempipacman}~{\em (c)} reports total execution times for all the four possible combinations grounder+solver.
All in all, we can conclude that the worse performance in solving times is widely absorbed by the time savings due to the use of incremental grounding.

As shown in Figure~\ref{fig:memoriapacman}~{\em (a)}, the overgrounding engine stores progressively more rules.
The impact of larger inputs for solvers is almost constant, while incremental grounding times have a slight increase on later iterations.
We can notice that the number of added rules per each iteration follows a steep initial curve and almost stabilizes in later iterations.
Nonetheless, almost all useful rules are added in the first iteration.
Figure~\ref{fig:memoriapacman}~{\em (b)} reports the recorded peak memory for the overgrounding engine when run with an horizon of $10,20$ and $30$ tiles respectively.

%%%%%%%%%%%%%%%%%%%%%%%%%%%%%%%%%%%%%%%%%%%%%%%%%%%%%%%%%%%%%%%%%%%%%%%%%%%
\section{Related Work}
Although we focus more on time requirements rather than on dealing with huge amounts of streamed data,
our proposal is comparable to early and recent work on incremental update of Datalog materializations~\cite{DBLP:journals/ai/MotikNPH19}.
In such work, {\em incremental maintainance} refers to a setting in which a pool of views (i.e. a logic program/knowledge base) is run repeatedly over a changing database. The views are updated from a run to another, and both insertion or deletion of tuples in input data trigger update activities on views.
In our paper we adopt the same terminology, and we refer to a logic program $P$ and to facts $F_1, \dots, F_n$. $AS(P \cup F_1)$ can be, in principle, totally different from $AS(P \cup F_2)$.
However, $AS(P \cup F_2)$ can be computed on $Inst(P , F_1 \cup F_2)^\infty$ which includes all the rules of $Inst(P, F_1)^\infty$, were this latter should be already pre-computed, thus reducing grounding times.
However, the aforementioned approaches focus on query answering over stratified Datalog programs, and can just materialize query answers; our focus is on a generalized setting, where disjunction and unstratified negation are allowed and propositional logic programs are materialized and maintained.
Furthermore, we purposely decided to not introduce deletion techniques when logic assertions are retracted between two consecutive runs: our overgrounded programs, while keeping semantic soundness, grow monotonically; yet, the absence of negative update activities allows to keep incremental grounding times significantly low.

Our work has also connections with the iclingo/oclingo approach~\cite{DBLP:conf/lpnmr/GebserKKS11,DBLP:journals/tplp/GebserKKS19} where incrementality is meant in a slightly different way: one has a base program $P$, which is coupled
with additional module layers $M_1, \dots, M_n$.
An answer set $A$ of $P \cup M_1$ can be \quo{incremented} with new atoms so to build an answer set $A'$ of $P \cup M_1 \cup M_2$.
%Note that, as implicitly observed by Reviewer 2, $A$ cannot be changed or be updated/recomputed, but it will be a subset of $A'$.
In other words, according to the iclingo/oclingo philosophy, one has to model her/his problem by thinking in terms of \quo{layers} of modules.

To date, incremental solving is (still) an almost unexplored topic. By full incremental solving, we mean the widest general settings, in which both $P$ and $F$ are subject to unexpected changes.
This research line would require to go beyond both our overgrounding-based and iclingo/oclingo approaches, by removing annotations and constraints in modelling, and introducing seamless updates of answer sets without restarting solving and/or grounding.
The Ticker system~\cite{DBLP:journals/tplp/BeckEB17} can be seen as a significant effort in this direction as it implements the LARS stream reasoning formal framework by using truth maintenance techniques, under ASP semantics.

In our work, we left full incremental reasoning out of the table, as the focus of the paper is on grounding-intensive repeated tasks. We note that such tasks are not necessarily polynomial, as unstratification and disjunction is possible.

Furthermore, it is worth noting that overgrounding is essentially orthogonal to recent advances in lazy grounding~\cite{DBLP:journals/fuin/PaluDPR09,DBLP:journals/tplp/LefevreBSG17,DBLP:conf/ijcai/BogaertsW18}; indeed, these essentially aim at blending grounding tasks within the solving step for reducing memory consumption; rather, our focus is on making grounding times negligible on repeated evaluations by explicitly allowing the usage of more memory, yet still keeping the two evaluation steps separated.

\vspace{-1em}
\section{Conclusions}
In this work we reported about theoretical properties of embedded and overgrounded programs, and the consequent development of an incremental grounder with permanent in-memory storage of ground programs.
Experiments show the potential of the approach; it must be noted that our ground program caching strategy is remarkably simple: cached ground programs grow monotonically between consecutive shots, thus becoming progressively larger but more generally applicable to a wider class of sets of input facts.
We measured a considerable decrease in grounding times and in the number of newly added rules in later shots; interestingly, the impact of larger ground instances on model generators is fair.
All in all, the results of benchmarks regarding memory footprints and the time performance clearly qualify the overgrounding approach for applications in which speed requirements are very tight and can be reached affording more memory.

The simplicity of the overgrounding approach paves the way to several extensions.
We are currently investigating: the possibility of discarding rules when a memory limit is required;  the impact of updates (i.e., additions/deletions of rules) in selected parts of the logic program; the introduction of ground programs which keep the properties of embeddings, yet allowing some form of simplification policies.
Further experiments, benchmark encodings and the binary of \system are available at~{\small \url{https://www.mat.unical.it/calimeri/material/iclp2019}}. %~\cite{incrwebsite}.

\section*{Acknowledgements}
This work has been partially supported by the Italian MIUR Ministry and the Presidency of the
Council of Ministers under project \quo{Declarative Reasoning over
Streams} under the \quo{PRIN} 2017 call (CUP $H24I17000080001$, project
2017M9C25L\_001), by the Italian MISE Ministry under project \quo{S2BDW} (F/050389/01-03/X32) - \quo{Horizon2020} PON I\&C2014-20
and by Regione Calabria under project \quo{DLV LargeScale} (CUP $J28C17000220006$) -
POR 2014-20.

\bibliographystyle{acmtranc}
%\bibliography{references}

\newpage
\appendix

\section{Proofs for Section~\ref{sec:embedding} and Section~\ref{sec:overgrounding}}
%\todo{la label 'appendix a.1' \`{e} bruttissima}
\label{appendix}

\begin{lemma }\label{lem:stageTorule}\rm
It is given a program $P$, a set $F$ of facts, an embedding program $\EP{}$ of $P \cup F$ and an answer set $A \in AS(P \cup F)$. Then, for each $a \in A$ there exists a rule $r_a \in (\grnd(P) \cup F)$ s.t. $a \in H(r_a)$ and $r_a \in \EP{}$; thus, $A \subseteq \HA(\EP{})$.
\end{lemma }

\begin{proof}
By Theorem~\ref{theo:therearestages}, each $a \in A$ is associated to an integer value $stage(a)$ and there exists a rule $r_a \in grnd(P) \cup F$, with $a \in H(r_a)$. Note that $r_a \in (grnd(P) \cup F)^A$ since $A \models B(r)$.
We now show that $r_a \in \EP{}$\xspace by induction on the {\em stage} associated to $a \in A$.
If $stage(a)=1$, $r_a$ is such that $B(r_a) = \emptyset$.
Hence, since $\EP{}$ is an embedding program for $P \cup F$, and $\EP{} \embeds_b r_a$, it must hold that $r_a \in \EP{}$.
Now, (inductive hypothesis) assume that for $stage(a) < j$, $r_a \in \EP{}$.
We show that for $stage(a)= j$, $r_a \in \EP{}$.
Indeed $r_a$ is such that for each $b \in B^+(r_a)$, $stage(b)< j$, and hence there exists a rule $r_b \in \EP{}$ with $b \in H(r_b)$.
Hence $\EP{} \embeds_b r_a$ and thus, since $\EP{}$ is an embedding program for $P \cup F$, $\EP{} \embeds_h r_a$, and $r_a \in \EP{}$.
\end{proof}

\medskip

%\begin{proof}
\noindent {\em Proof of} {\em Theorem~\ref{theo:Embequivalence}}
 
\noindent [{\bf $AS(\grnd(P) \cup F) \subseteq AS(\EP{})$}].\ \ \ Let $A \in AS(\grnd(P) \cup F)$.
We will show that $(\grnd(P) \cup F)^A = \EP{}^A$, thus the statement trivially follows.
Indeed, since $\EP{} \subseteq \grnd(P) \cup F$, it holds that $\EP{}^A \subseteq (\grnd(P) \cup F)^A$.
So, if $(\grnd(P) \cup F)^A$ and $\EP{}^A$ differ, there must exists a rule $r \in \grnd(P) \setminus \EP{}$, and, obviously,
such that $r \in (\grnd(P) \cup F)^A$.
But since $\EP{}$ is an embedding program for $P \cup F$, this
means that $\EP{} \nembeds B(r)$.
However, $A \models B(r)$, and hence $\forall b \in B^+(r)$ we
know that $b \in A$.
By Lemma~\ref{lem:stageTorule}, $A \subseteq \HA(\EP{})$,
and then $\forall b \in B^+(r)$ there exists a rule $r' \in \EP{}$ such that $b \in H(r')$, thus leading to a contradiction with $\EP{} \nembeds B(r)$.

\noindent [{\bf $AS(\EP{}) \subseteq AS(\grnd(P) \cup F)$}].\ \ \ Let $A \in AS(\EP{})$. We again show that $\EP{}^A = (\grnd(P) \cup F)^A$.
Similarly to the case above, since $\EP{}^A \subseteq (\grnd(P) \cup F)^A$, there must exists a rule $r \in (\grnd(P) \cup F) \setminus \EP{}$, s.t. $\EP{} \nembeds B(r)$. Moreover, $A \models B(r)$, and thus
$\forall b \in B^+(r)$ we know that $b \in A$.
However, $A$ is an answer set for $\EP{}$; this clearly means that $\forall b \in B^+(r)$ there exists $r' \in \EP{}$ such that $b \in H(r')$, which in turn means that $\EP{} \embeds B(r)$, thus leading to a contradiction.
%\end{proof}

\medskip

%\begin{proof}
\noindent {\em Proof of} {\em Proposition~\ref{prop:IntersectionEP}}

\noindent By contradiction, assume that $\EP{}$ is not an embedding program for $P \cup F$.
Then, $\exists r \in (\grnd(P) \cup F)$ such that $\EP{} \nembeds r$, that is, $\EP{} \nembeds_h r$ and $\EP{} \embeds_b r$.
Since $\EP{} \nembeds_h r$, we have that $r \notin \EP{}$, and hence at least one of the following statements hold: $(i)$ $r \notin \EP{1}$, $(ii)$ $r \notin \EP{2}$.
Without loss of generality, assume $r \notin \EP{1}$.
By hypothesis, $\EP{1}$ is an embedding program for $P \cup F$, thus it must hold that $\EP{1} \nembeds_b r$.
Then, by definition, there exists $b \in B^+(r)$ s.t. $\nexists r' \in \EP{1}$ with $b \in H(r')$; this implies that such $r'$ cannot exist in $\EP{}$, thus contradicting the fact that $\EP{} \embeds_b r$.
%\end{proof}

\medskip

%\begin{proof}
\noindent {\em Proof of} {\em Theorem~\ref{theo:fixedpointsareEPs}}

\noindent ($\Rightarrow$) Assume that $\EP{}$ is an embedding program for $P \cup F$.
By contradiction, assume there is a rule $r \in \instt{P}{\EP{}}\cup F$ such that  $r \notin \EP{}$.
Clearly, $B^+(r) \subseteq Heads(\EP{})$ (by definition of $\inst$).
This means that $\EP{} \embeds_b r$, and, since $\EP{} \embeds r$, this implies $\EP{} \embeds_h r$, i.e., $r \in \EP{}$, thus contradicting our assumption.

\noindent ($\Leftarrow$) Assume that for a set of rules $\EP{}$, $\EP{} \supseteq \instt{P}{\EP{}} \cup F$ and,
by contradiction, that $\EP{}$ is not an embedding program for $P \cup F$.
Then, there must be a rule $r \in \grnd(P) \cup F$ such that $\EP{} \nembeds r$.
Clearly, $r \notin \EP{}$ (otherwise, we would have $\EP{} \embeds_h r$ ), and $\EP{} \embeds_b r$.
This means that $B^+(r) \subseteq Heads(\EP{})$ and thus $\instt{P}{\EP{}} \subseteq \EP{}$ must contain $r$, contradicting our assumption.
%\end{proof}

\medskip

%\begin{proof}
\noindent {\em Proof of} {\em Theorem~\ref{cor:intersectionOfEmbedding}}

\noindent Let define the monotone operator ${\overline{\inst}}(P,F) = \inst(P,F) \cup F$, and consider the complete lattice of subsets of $\grnd(P) \cup F$ under set containment.

The proof then follows by Theorem~\ref{theo:fixedpointsareEPs} and by Knaster-Tarski theorem~\cite{tarski1955} by observing that
\[ \insttinf{P}{F}\cup F = l\!f\!p({\overline{\inst}}) = in\!f \{ \EP{} \subseteq \grnd(P) \cup F \mid {\overline{\inst}}(P,F) \subseteq \EP{} \} =  \bigcap_{\EP{} \in \EPS} \EP{} \] and that $\insttinf{P}{F}\cup F$ is the least fixpoint for ${\overline{\inst}}(P,F)$.
%\end{proof}

\medskip

%\begin{proof}
\noindent {\em Proof of} {\em Theorem~\ref{theo:intersequivalence}}

\noindent The proof follows from Theorem~\ref{theo:Embequivalence} and Proposition~\ref{prop:IntersectionEP}.
% \end{proof}

\medskip

%\begin{proof}
\noindent {\em Proof of} {\em Theorem~\ref{theo:core}} 

\noindent Let $IU = \insttinf{P}{{\UF}_k}\cup F_i$, and  for each $i$, $i \leq k$, $IF_i = \insttinf{P}{F_i}\cup F_i$. Recall that each $IF_i$ is an embedding for $P \cup F_i$.

By monotonicity of $\inst$, we have point~(1) above and that $IU \supseteq IF_i$ for each $i \leq k$. Each $IF_i$ is clearly an embedding program for $P \cup F_i$ by Theorem~\ref{cor:intersectionOfEmbedding}.
We show that point~2 follows by showing that $IU$ is an embedding program for $P \cup F_i$ and from Theorem~\ref{theo:Embequivalence}.

For a given $i \leq k$, consider a rule $r \in (\grnd(P)\cup F_i)$;\ If $r \in IU$ then $IU \embeds r$.
Let us consider the case in which $r \in (\grnd(P)\cup F_i) \setminus IU$.
Note that $IF_i \embeds r$ and thus $IF_i \nembeds_b r$.
Now, either $IU \nembeds_b r$ or $IU \embeds_b r$.
In the former case, clearly $IU \embeds r$.
In the latter case, we have that $\forall a \in B^+(r)$ $a \in Heads(IU)$, and this means that $r \in IU$ by definition of $IU$, thus contradicting the assumption that $r \not\in IU$.
Thus $IU$ is an embedding for $P \cup F_i$ for all $i \leq k$.
%\end{proof}

%\newpage
%\input{answerReviewers.tex}
\end{document}